%% file: paper.tex
\documentclass[]{gensi}
\usepackage{tabularx}
\usepackage[toc,page,header]{appendix}

\usepackage{minitoc}
\usepackage{cleveref} 
\usepackage{microtype}
\usepackage{subcaption}
\usepackage{amsmath} 
\usepackage{amssymb} 
\usepackage{amsfonts}       
\usepackage{pgfplots}
\usepackage{pgfplotstable}
\usepackage{xcolor}\usepackage{fontawesome5}
\usepackage{CJKutf8}
\usetikzlibrary{patterns}
\usepackage{multirow}
\usepackage{setspace}
\usepackage{hyperref}
\tcbuselibrary{theorems}
\usepackage{float}
\usepackage{caption}
\usepackage{wrapfig}
\usepackage{xspace}
\usepackage{graphicx}      
\usepackage{subcaption}    
\usepackage{seqsplit}      

\usepackage{siunitx}
\usepackage{soul}          
\usepackage{array}
\usepackage{pifont}

\usepackage{cleveref}
\usepackage[table]{xcolor}

\usepackage{tcolorbox}
\newtcolorbox{DefinitionBox}{
  colback=blue!5,
  colframe=blue!80,
  boxrule=0.5pt,
  arc=2pt,
  left=2pt,
  right=2pt,
  top=2pt,
  bottom=2pt,
}

\newtcolorbox{CorollaryBox}{
  colback=gray!5,
  colframe=gray!80,
  boxrule=0.5pt,
  arc=2pt,
  left=2pt,
  right=2pt,
  top=2pt,
  bottom=2pt,
}

\usepackage{listings}
\definecolor{promptbackground}{HTML}{F7F8FA}
\definecolor{promptframe}{HTML}{C8CED8}
\definecolor{plmcolor}{HTML}{FCE4D6}    
\definecolor{llmcolor}{HTML}{D9EAF7}    
\definecolor{agentcolor}{HTML}{E2F0D9}  

\lstdefinestyle{promptstyle}{
    basicstyle=\ttfamily\scriptsize,
    breaklines=true,
    breakatwhitespace=false,
    breakautoindent=false,
    breakindent=0pt,
    postbreak=\mbox{},
    columns=fullflexible,
    keepspaces=true,
    showstringspaces=false,
    frame=none
}

\newtcblisting{promptbox}[1]{
    enhanced,
    breakable,
    listing only,
    title={#1},
    colback=promptbackground,
    colframe=promptframe,
    coltitle=black,
    fonttitle=\bfseries\small,
    boxrule=0.4pt,
    arc=1.2mm,
    left=2mm,
    right=2mm,
    top=1mm,
    bottom=1mm,
    toptitle=1.6mm,
    bottomtitle=1mm,
    listing options={
        style=promptstyle,
        moredelim={[is][\color{red}\bfseries]{<<}{>>}}
    }
}

\usepackage{fix-cm}

\input{macro}
\input{math_commands}

\newcommand{\bench}{PFArena\xspace}

\newcommand{\toolkit}{AMixKit\xspace}

\newcommand{\ntaskone}{\texttt{T1:\allowbreak\hspace{0.2em}Single-{\allowbreak}mutant\allowbreak\hspace{0.35em}generation}\xspace}
\newcommand{\ntasktwo}{\texttt{T2:\allowbreak\hspace{0.2em}Measurement-{\allowbreak}free\allowbreak\hspace{0.35em}multi-{\allowbreak}mutant\allowbreak\hspace{0.35em}ranking}\xspace}
\newcommand{\ntaskthree}{\texttt{T3:\allowbreak\hspace{0.2em}Anchor-{\allowbreak}informed\allowbreak\hspace{0.35em}multi-{\allowbreak}mutant\allowbreak\hspace{0.35em}ranking}\xspace}
\newcommand{\ntaskfour}{\texttt{T4:\allowbreak\hspace{0.2em}Single-{\allowbreak}mutant-{\allowbreak}informed\allowbreak\hspace{0.35em}multi-
{\allowbreak}mutant\allowbreak\hspace{0.35em}ranking}\xspace}
\newcommand{\ntone}{\texttt{T1}\xspace}
\newcommand{\nttwo}{\texttt{T2}\xspace}
\newcommand{\ntthree}{\texttt{T3}\xspace}
\newcommand{\ntfour}{\texttt{T4}\xspace}
\newcommand{\dmsscore}{\texttt{DMS\_score}\xspace}

\definecolor{myclr}{RGB}{201, 236, 255}

\title{
PFArena: Benchmarking Language Models for Protein Modification
}

\affiliation{
Shanghai Artificial Intelligence Laboratory\\
Generative Symbolic Intelligence Lab (GenSI), Tsinghua University\\
Institute for AI Industry Research (AIR), Tsinghua University \\
School of Pharmaceutical Sciences, Tsinghua University
}

\newcommand{\projectlinks}{\hspace{0.5em}\href{https://github.com/AMix-Bio/PFArena}{\textcolor{black}{\faGithub\ Github}}\hspace{2.5em}\href{https://huggingface.co/datasets/AMix-Bio/PFArena}{\textcolor{black}{\faDatabase\ Dataset}}\hspace{2.0em}\href{https://huggingface.co/spaces/AMix-Bio/PFArena-Leaderboard}{\textcolor{black}{\faTrophy\ Leaderboard}}}

\providecommand{\abstractlist}{}
\renewcommand{\beginabstract}{}
\begin{document}

    \maketitle
    \vspace{-0.8em}

    \begin{figure}[H]
        \centering
        \includegraphics[width=1.0\linewidth]{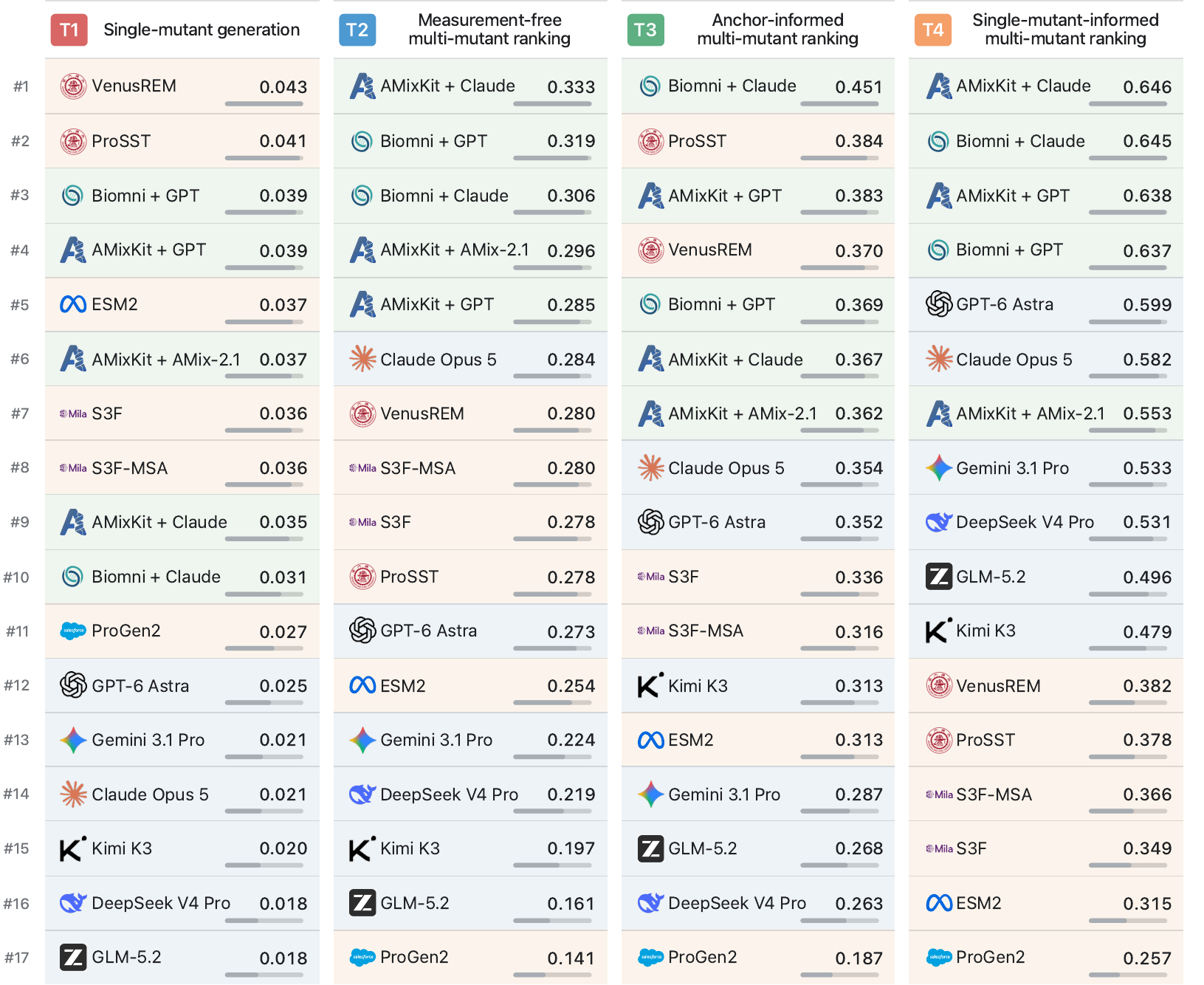}
        \caption{\textbf{Benchmark performance across the four protein modification tasks in \bench.}
Method classes are distinguished by cell background color:
{\textcolor{plmcolor}{\rule{0.7em}{0.7em}}} protein language models,
{\textcolor{llmcolor}{\rule{0.7em}{0.7em}}} large language models, and
{\textcolor{agentcolor}{\rule{0.7em}{0.7em}}} agents.
For agents, Claude denotes Claude Opus 5 and GPT denotes GPT-6 Astra.
Methods are ranked by Recall@$40$ for task \ntone and by Spearman correlation for tasks \nttwo--\ntfour.
Results for the remaining metrics are detailed in the experiments section.
}
        \label{fig:overview}
    \end{figure}
\footnotetext{Correspondence: \href{mailto:zhouhao@air.tsinghua.edu.cn}{zhouhao@air.tsinghua.edu.cn}}

    \newpage
    \tableofcontents
    \newpage

    \section*{Abstract}
\input{sections/00abstract}
    \input{sections/01intro}

    \input{sections/03PFArena}
    
    \input{sections/04experiments}

    \input{sections/05analysis}

    \input{sections/02related}
    
    \input{sections/06conclusion}

    \input{sections/contributions}
    
    \clearpage
    \bibliographystyle{unsrtnat}
    \bibliography{refs}

    \clearpage
    \beginappendix
    \input{sections/appendix}

\end{document}

%% file: macro.tex
\usepackage{natbib}
\usepackage{latexsym}

\usepackage{url}
\usepackage{amssymb}
\usepackage[utf8]{inputenc}
\usepackage{microtype}
\usepackage{booktabs}
\usepackage{pifont} 
\usepackage{multirow}
\usepackage{makecell}
\usepackage{paralist}
\usepackage{xspace}
\usepackage{color}
\usepackage{xcolor}
\usepackage{colortbl}
\usepackage{adjustbox}
\usepackage{hyperref} 
\usepackage[edges]{forest}
\usepackage{tikz} 
\usepackage{caption}
\usepackage{amsfonts}
\usepackage{tcolorbox}
\usepackage{algorithm}
\usepackage{algpseudocode}
\usepackage{amsthm}

\usepackage{mathtools}
\usepackage[version=4]{mhchem}
\usepackage{array}  
\usepackage{graphicx}

\hypersetup{
    colorlinks,
    linkcolor={blue!80!black},
    citecolor={blue!80!black},
}
\tikzset{
    root/.style =             {align=center, text width=1cm, rounded corners=3pt, line width=0.3mm, fill=gray!10, draw=gray!80, font=\small},
    demographic/.style =         {align=center, text width=1.8cm, rounded corners=3pt, line width=0.3mm, fill=blue!10, draw=blue!80, font=\footnotesize},
    demographic_work/.style =    {align=center, text width=10cm, rounded corners=3pt, line width=0.3mm, fill=blue!10, draw=blue!0, font=\footnotesize},
    character/.style =         {align=center, text width=1.8cm, rounded corners=3pt, line width=0.3mm, fill=red!10, draw=red!80, font=\footnotesize},
    character_work/.style =    {align=center, text width=10cm, rounded corners=3pt, line width=0.3mm, fill=red!10, draw=red!0, font=\footnotesize},
    personalization/.style =           {align=center, text width=1.8cm, rounded corners=3pt, line width=0.3mm, fill=cyan!10, draw=cyan!80, font=\footnotesize},
    personalization_work/.style =      {align=center, text width=10cm, rounded corners=3pt, line width=0.3mm, fill=cyan!10, draw=cyan!0, font=\footnotesize},
    risk/.style =         {align=center, text width=1.8cm, rounded corners=3pt, line width=0.3mm, fill=orange!10, draw=orange!80, font=\footnotesize},
    risk_work/.style =    {align=center, text width=10cm, rounded corners=3pt, line width=0.3mm, fill=orange!10, draw=orange!0, font=\footnotesize},
}

\usepackage{CJK}


%% file: math_commands.tex
\usepackage{amsmath,amsfonts,bm}

\def\eqref#1{equation~\ref{#1}}

\def\1{\bm{1}}

\DeclareMathAlphabet{\mathsfit}{\encodingdefault}{\sfdefault}{m}{sl}
\SetMathAlphabet{\mathsfit}{bold}{\encodingdefault}{\sfdefault}{bx}{n}



%% file: sections/00abstract.tex
Protein modification requires navigating an immense sequence space, yet wet-lab validation remains low-throughput and costly. Although computational paradigms including protein language models (PLMs), large language models (LLMs), and LLM-based agents have shown promise in protein modification, their relative efficacy across realistic experimental decision-making settings remains unclear. 
To bridge this gap, we introduce \textbf{\bench}, a benchmark comprising four controlled task interfaces that cover single-mutant generation and multi-mutant ranking. 
By providing varying levels of mutation fitness data, \textbf{\bench} reflects four representative research scenarios characterized by differing degrees of prior experimental context. 
We assess six PLMs, six LLMs, and five LLM-based agents using complementary metrics to measure both peak and overall protein modification performance. 
Our evaluation reveals that model performance shifts systematically with the availability of target-specific experimental evidence: 
PLMs demonstrate proficiency in open-ended single-mutant generation by leveraging protein-specific priors, whereas LLMs and agents perform strongly in multi-mutant ranking, particularly when target-specific fitness data are available.
Nevertheless, all model families face fundamental challenges with the increase of search space and mutation depth. 
We release our code and benchmark suite to facilitate reproducible research in model-assisted protein modification.


%% file: sections/01intro.tex
\section{Introduction}

\label{sec:introduction}

Protein modification is central to protein engineering,
fundamentally relying on sequence mutations to alter protein properties.
This process results in an enormous number of possible substitutions and combinations,
making experimental characterization costly, time-consuming, and limited in throughput.
Reliable fitness predictions allow researchers to prioritize mutants with the highest expected value
and focus measurements on informative regions of the sequence–fitness landscape,
thereby accelerating iterative design–build–test-learn cycles
and supporting the discovery of mutations difficult to identify through empirical screening alone.

Computational models, including protein language models (PLMs) and large language models (LLMs), have shown substantial potential for protein modification. PLMs learn sequence, structural, and evolutionary constraints from large-scale protein data, enabling zero-shot prediction and prioritization of functional mutants~\cite{lin2023evolutionary,nijkamp2023progen2,li2024prosst,zhang2024s3f,tan2025venusrem,frazer2021eve}. 
General-purpose LLMs have also advanced rapidly on protein-specific prediction tasks. On ProteinGym Hard, the reported score increased from 37.7\% for Claude Opus 4.7 to 39.6\% for Claude Opus 4.8, followed by a further 7.7-percentage-point gain for Claude Opus 5~\cite{anthropic2026claude48,anthropic2026claude5}. This sustained improvement highlights the growing ability of LLMs to interpret protein sequences through natural-language interaction. Nevertheless, it remains unclear when specialized PLMs, general-purpose LLMs, or LLM-based agents are preferable across the diverse decision settings encountered in protein modification.

To answer this question, we introduce \textbf{\bench}, an assay-grounded benchmark that systematically evaluates these model families under four protein modification settings: \ntaskone, \ntasktwo, \ntaskthree and \ntaskfour.
These settings vary along two axes: the action space and the available experimental evidence. They encompass both open-ended generation and fixed-pool ranking tasks, and cover conditions with different mutant information supplied. This design enables a systematic comparison between PLMs and LLMs across single- and multi-mutant prioritization, while testing whether LLMs can translate target-specific experimental evidence into better mutation ranking.

We evaluate six PLMs and six LLMs across four tasks covering 202 unique assays and 293 task instances. Furthermore, to combine the reasoning capabilities of LLMs with protein-specific tools, we evaluate five LLM-based agents built on both the pioneering Biomni framework~\cite{huang2025biomni} and our lightweight \textbf{\toolkit} toolkit, detailed in \Cref{sec:implementation-details}. 
Our study yields two main findings: 

\begin{itemize}
    \item \textbf{Divergent strengths.}
    No single model family dominates across all four tasks. PLMs hold an overall advantage in the open-ended \ntone setting, where no target-specific fitness measurements are available and performance depends largely on protein-specific sequence, structural, and evolutionary priors. Their predictions also span a broader range of residue-class substitutions. In contrast, LLM-based systems perform strongly in the multi-mutant ranking tasks (\nttwo--\ntfour). Their advantage is most evident when comprehensive target-specific experimental evidence is provided (\ntfour). Together, these results highlight the complementary strengths of protein-specific representations and evidence-guided reasoning.

    \item \textbf{Convergent bottlenecks.}
    Despite these distinct strengths, all three model families encounter shared limitations. In \ntone, even the best-performing method recovers fewer than 5\% of the ground-truth top-$40$ mutations, showing that the single-mutant search space remains difficult to cover under a limited prediction budget. Across \nttwo--\ntfour, all model families show a substantial drop in ranking accuracy beyond two substitutions, revealing a shared sensitivity to mutation depth.
\end{itemize}

%% file: sections/03PFArena.tex
\section{\bench}
\label{sec:benchmark}

\begin{figure*}[t]
      \centering
      \includegraphics[width=\textwidth]{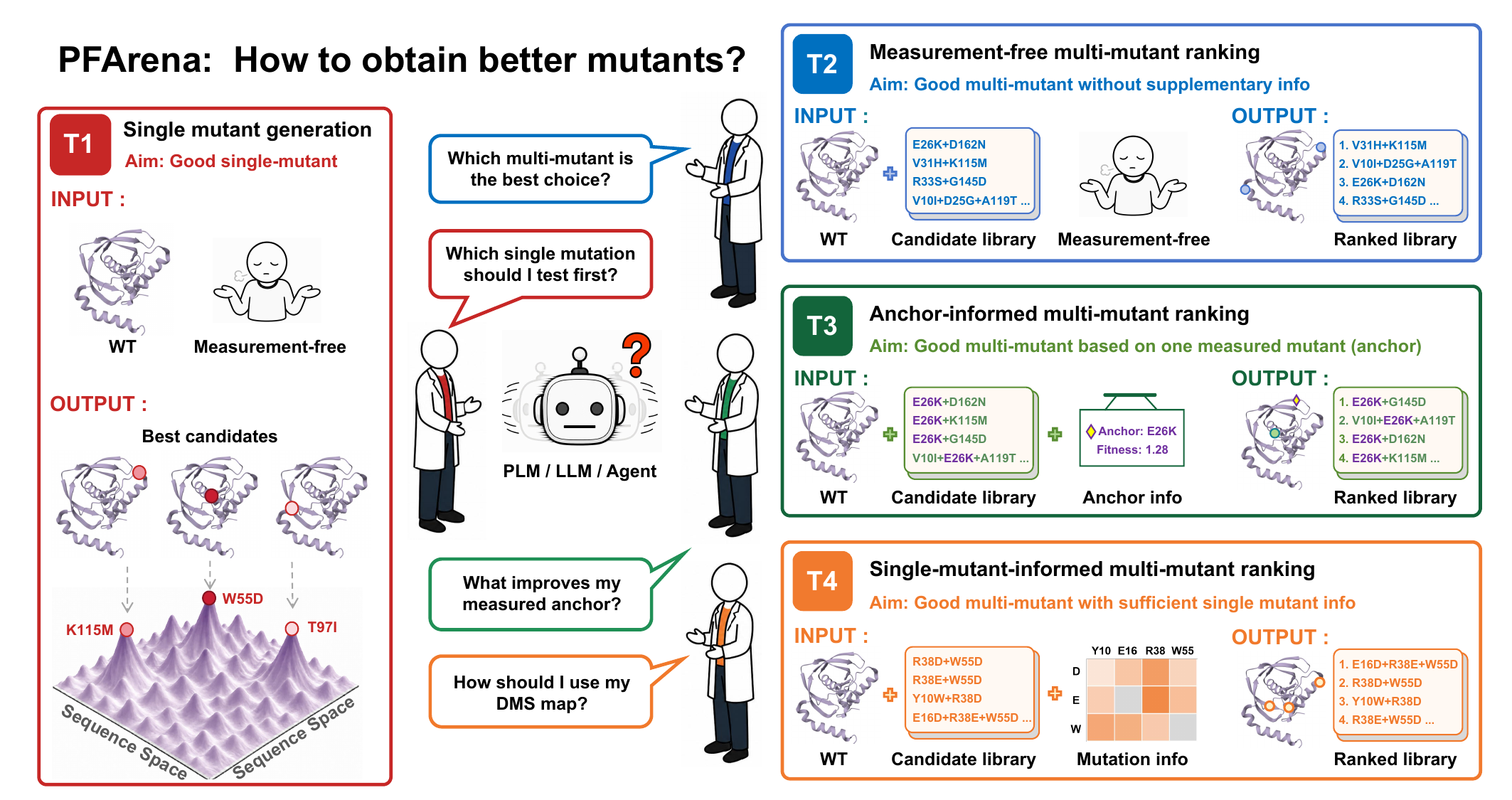}
      \caption{\textbf{PFArena evaluates protein modification from the perspective of experimental scientists.}
      The benchmark is organized around four representative research scenarios that correspond to the questions scientists ask when deciding which mutants to study next. 
      \textbf{T1: Single-mutant generation} generates high-performing single mutations without candidate input.
      \textbf{T2: Measurement-free multi-mutant ranking} ranks multi-mutant candidates without prior fitness measurements.
      \textbf{T3: Anchor-informed multi-mutant ranking} ranks multi-mutant successors conditioned on a tested anchor mutant. 
      \textbf{T4: Single-mutant-informed multi-mutant ranking} ranks multi-mutant candidates using measured single-mutant fitness information. These scenarios span single- and multi-mutant discovery under progressively richer experimental evidence, enabling systematic comparison of PLMs, LLMs, and agents in realistic protein-engineering workflows.
 The protein structure depicted here corresponds to PDB entry 1RQC \cite{robien2004improved}.
      }
      \label{fig:pfarena}
  \end{figure*}

Conventional mutant-effect benchmarks typically assess whether a model can assign fitness scores to prespecified mutations~\cite{notin2023proteingym,dallago2021flip}. In contrast, \textbf{\bench} evaluates whether PLMs, LLMs, and agents can prioritize mutants across four task interfaces. These interfaces vary in candidate space, spanning single substitutions and mutation combinations, and the amount and form of target-specific evidence available in each task.

\subsection{Task Suite}
\label{sec:task-suite}

As summarized in \Cref{fig:pfarena}, \textbf{\bench} defines four task interfaces that correspond to common decisions in protein-engineering workflows. \ntone considers the open-ended generation of single-mutant candidates from the full legal substitution space, whereas \nttwo--\ntfour consider the ranking of a supplied pool of multi-mutant candidates under different levels of target-specific experimental evidence. 

Formal task definitions are provided in the following subsections, and the specification tables summarize the information available at the benchmark level. Because model families expose different native interfaces, this information is presented through model-appropriate interfaces. Natural-language task inputs are provided directly to LLMs and agents, whereas PLMs receive the subset supported by their architectures.

\subsubsection{T1: Single-mutant generation}
\label{sec:task-t1}

\begin{table*}[t]
\centering
\small
\vspace{0.5em}
\renewcommand{\arraystretch}{1.2}
\begin{tabular}{@{}p{0.18\textwidth} p{0.74\textwidth}@{}}
\toprule
\addlinespace[0.5em]
\textbf{Task component} & \textbf{Specification} \\
\addlinespace[0.2em]
\midrule
Definition & Budgeted generation of legal single substitutions without target-specific mutation measurements. \\
\midrule
Input & Wild-type sequence $x$, assay context $c$ and proposed budget $K$. \\
\midrule
Output & Proposed mutation set of at most $K$ single substitutions. \\
\midrule
Instance & \textbf{Input:}\newline 
\hspace*{1.5em}Wild-type sequence: \texttt{MSSSLGKE...AMV}\newline 
\hspace*{1.5em}Assay context: \newline
    \hspace*{1.5em} - Primary task class: activity function. \newline
    \hspace*{1.5em} - Fitness type: organismal or cellular fitness. \newline
    \hspace*{1.5em} - Readout subclass: growth or selection proxy. \newline 
\hspace*{1.5em}Proposed budget $K$: 40 \vspace{0.3em}\newline
\textbf{Output:}\newline 
\hspace*{1.5em}Proposed mutation set: \texttt{P557L}, \texttt{K369Y}, \texttt{N434D}, \texttt{...}. \\
\midrule
Statistics & Assays: 123; measured single mutants: 594,695. \\
\bottomrule
\end{tabular}
\caption{Task specification for \ntaskone.}
\label{tab:task-t1}
\end{table*}

In a single-mutant campaign like deep mutational scanning (DMS), researchers assay a broad panel of legal single substitutions around a wild-type protein and use the measurements to identify mutants with favorable assay-specific phenotypes. Task \ntone evaluates the analogous budgeted prioritization problem: given the wild-type sequence and assay context but no target-specific mutation measurements, which single substitutions should be selected first under a fixed experimental budget?

\Cref{tab:task-t1} summarizes the task specification. Formally, given a wild-type sequence $x$, assay context $c$, and a generation budget $K$, the model implements a generation function:

\[
f_{\theta}:(x,c,K)\mapsto\widehat{\mathcal{O}},
\qquad
\widehat{\mathcal{O}}\subseteq\mathcal{V}(x),
\qquad
|\widehat{\mathcal{O}}|\leq K,
\]

where $\mathcal{V}(x)$ denotes the legal single-substitution space of a canonical wild-type sequence $x$. For a sequence of length $L$, this space contains $19L$ possible substitutions under the standard amino-acid alphabet. The model must therefore return a limited set of legal candidates rather than a complete ordering of the entire space.

\ntone assesses highly selective discovery in a large single-mutant space under a finite experimental budget. A successful system should identify at least one highly promising mutation while also exploring a sufficiently broad region of the high-fitness landscape. For smaller PLMs, the task can be approached by evaluating all legal substitutions individually and ranking their scores. For LLMs and agents, exhaustive evaluation is generally impractical because of the associated computational and financial costs, so these systems must directly generate a small candidate set from the full substitution space.

Single-mutant discovery provides a natural starting point for protein engineering, but many desired phenotypes depend on combinations of substitutions that may reinforce, compensate for, or interfere with one another. The remaining tasks therefore move from open-ended single-mutant generation to fixed-pool ranking of multi-mutant candidates with progressively richer levels of target-specific experimental evidence made available.

\subsubsection{T2: Measurement-free multi-mutant ranking}
\label{sec:task-t2}

\begin{table*}[t]
\centering
\small
\vspace{0.5em}
\renewcommand{\arraystretch}{1.2}
\begin{tabular}{@{}p{0.18\textwidth} p{0.74\textwidth}@{}}
\toprule
\addlinespace[0.5em]
\textbf{Task component} & \textbf{Specification} \\
\addlinespace[0.2em]
\midrule
Definition & Ranking multi-mutant candidates without target-specific mutation measurements. \\
\midrule
Input & Wild-type sequence $x$, assay context $c$, and a score-blind candidate pool. \\
\midrule
Output & Complete ranking of the supplied candidate pool. \\
\midrule
Instance & \textbf{Input:}\newline 
\hspace*{1.5em}Wild-type sequence: \texttt{MLEGKVKW...KEA}\newline 
\hspace*{1.5em}Assay context: \newline
    \hspace*{1.5em} - Primary task class: stability. \newline
    \hspace*{1.5em} - Fitness type: stability. \newline
    \hspace*{1.5em} - Readout subclass: folding free energy stability readout. \newline 
\hspace*{1.5em}Candidate pool: \{\texttt{V28C+V63L}, \texttt{V28Q+V63P}, \texttt{V28M+V63F}, \texttt{...}\}. \vspace{0.3em}\newline 
\textbf{Output:}\newline 
\hspace*{1.5em}Ranked candidate set: \texttt{V28C+V63L}, \texttt{V28M+V63F}, \texttt{V28Q+V63P}, \texttt{...}. \\
\midrule
Statistics & Assays: 74; candidate number: 6,288. \\
\bottomrule
\end{tabular}
\caption{Task specification for \ntasktwo.}
\label{tab:task-t2}
\end{table*}

\begin{table*}[t]
\centering
\small
\vspace{0.5em}
\renewcommand{\arraystretch}{1.2}
\begin{tabular}{@{}p{0.18\textwidth} p{0.74\textwidth}@{}}
\toprule
\addlinespace[0.5em]
\textbf{Task component} & \textbf{Specification} \\
\addlinespace[0.2em]
\midrule
Definition & Ranking strict successors of a measured anchor mutant. \\
\midrule
Input & Wild-type sequence $x$, assay context $c$, a measured anchor mutant $u$ with score $y_u$, and a score-blind successor pool $\mathcal{C}$. \\
\midrule
Output & Complete ranking of the supplied candidate pool. \\
\midrule
Instance & \textbf{Input:}\newline 
\hspace*{1.5em}Wild-type sequence: \texttt{QVQLVQSG...VSS}\newline 
\hspace*{1.5em}Assay context: \newline
    \hspace*{1.5em} - Primary task class: binding. \newline
    \hspace*{1.5em} - Fitness type: binding. \newline
    \hspace*{1.5em} - Readout subclass: binding. \newline 
\hspace*{1.5em}Anchor: \texttt{T28P} with score -1.586. \newline 
\hspace*{1.5em}Successor pool: \newline \hspace*{1.5em} \{\texttt{T28P+S30R+N59K+T76A}, \texttt{T28P+S30R+N59K+Q62P+S75F}, \texttt{T28P+S30R+L104V}, \texttt{...}\}. \vspace{0.3em}\newline 
\textbf{Output:}\newline 
\hspace*{1.5em}Ranked candidate set: \newline \hspace*{1.5em} \texttt{T28P+S30R+N59K+Q62P+S75F}, \texttt{T28P+S30R+N59K+T76A}, \texttt{T28P+S30R+L104V}, \texttt{...}. \\
\midrule
Statistics & Assays: 67; candidate number: 3,648. \\
\bottomrule
\end{tabular}
\caption{Task specification for \ntaskthree.}
\label{tab:task-t3}
\end{table*}

Task \nttwo evaluates multi-mutant prioritization as fixed-pool ranking rather than open-ended generation. Given the wild-type sequence, assay context and a score-blind pool sampled from measured multi-mutants, the model returns a permutation of that pool for downstream experimental evaluation.

\Cref{tab:task-t2} summarizes the task specification. Formally, given a wild-type sequence $x$, assay context $c$, and a score-blind multi-mutant candidate pool $\mathcal{C}$, the model implements a ranking function:

\[
f_{\theta}:
\bigl(x,c,\mathcal{C}\bigr)
\mapsto
\widehat{\mathcal{O}},
\qquad
\widehat{\mathcal{O}}
\in
\Pi\bigl(\mathcal{C}\bigr).
\]

Here, $\Pi(\mathcal{C})$ denotes the set of all permutations of the candidate pool.

\nttwo assesses measurement-free combinatorial reasoning. Specifically, the model must prioritize variants using the available protein information and assay description while accounting for possible interactions among substitutions. It does not test whether a model can generate new combinations outside the supplied pool or assign calibrated fitness values to arbitrary mutants.

\subsubsection{T3: Anchor-informed multi-mutant ranking}
\label{sec:task-t3}

Task \ntthree evaluates successor ranking conditional on one measured sequence background. In the current release, 65 of 67 anchors are single substitutions; one anchor contains two substitutions and one contains three.

\Cref{tab:task-t3} summarizes the task specification. Formally, given a wild-type sequence $x$, assay context $c$, and a measured anchor mutant $u$ with score $y_u$, let $\mathcal{C}$ denote the supplied score-blind successor pool. Every candidate $v\in\mathcal{C}$ strictly contains the substitutions in $u$ and introduces at least one additional substitution relative to this measured background. The model then implements a ranking function:

\[
f_{\theta}:
\bigl(x,c,u,y_u,\mathcal{C}\bigr)
\mapsto
\widehat{\mathcal{O}},
\qquad
\widehat{\mathcal{O}}
\in
\Pi\bigl(\mathcal{C}\bigr).
\]

\ntthree assesses anchor-conditioned reasoning.
The measured anchor provides a local experimental reference; thus, the model must determine whether additional substitutions are likely to improve or reduce fitness relative to the measured sequence background.
This setting tests whether information from one experimentally characterized mutant can be transferred to the prioritization of related multi-mutant candidates.

\subsubsection{T4: Single-mutant-informed multi-mutant ranking}
\label{sec:task-t4}

Whereas \ntthree exposes one measured anchor, \ntfour exposes a broader set of measured single-mutant scores. This context is not required to cover the complete $19L$ single-substitution space. Instead, the construction requirement is candidate-specific completeness: every component substitution represented in a candidate has a corresponding measured single-mutant entry.

\Cref{tab:task-t4} summarizes the task specification. Formally, given a wild-type sequence $x$ and assay context $c$, let $\mathcal{S}$ map each provided measured single substitution $s$ to its assay score $y(s)$, and let $\mathcal{C}$ denote the supplied score-blind multi-mutant candidate pool, whose component substitutions are all represented in $\mathcal{S}$. The model then implements a ranking function:

\[
f_{\theta}:
\bigl(x,c,\mathcal{S},\mathcal{C}\bigr)
\mapsto
\widehat{\mathcal{O}},
\qquad
\widehat{\mathcal{O}}
\in
\Pi\bigl(\mathcal{C}\bigr).
\]

\ntfour assesses component-informed combination ranking. Given the measured effects of all individual mutations,
the model must prioritize multi-mutant combinations.
Rather than simply adding individual scores,
the system must account for non-additive interactions
to translate distributed evidence into accurate rankings.

\begin{table*}[t]
\centering
\small
\vspace{0.5em}
\renewcommand{\arraystretch}{1.2}
\begin{tabular}{@{}p{0.18\textwidth} p{0.74\textwidth}@{}}
\toprule
\addlinespace[0.5em]
\textbf{Task component} & \textbf{Specification} \\
\addlinespace[0.2em]
\midrule
Definition & Multi-mutant ranking with measured context for every component substitution. \\
\midrule
Input & Wild-type sequence $x$, assay context $c$, a measured single-mutant context map $\mathcal{S}$, and a score-blind multi-mutant candidate pool $\mathcal{C}$. \\
\midrule
Output & Complete ranking of the supplied candidate pool. \\
\midrule
Instance & \textbf{Input:}\newline 
\hspace*{1.5em}Wild-type sequence: \texttt{EVKLDETG...EIK}\newline 
\hspace*{1.5em}Assay context: \newline
    \hspace*{1.5em} - Primary task class: stability. \newline
    \hspace*{1.5em} - Fitness type: abundance or expression. \newline
    \hspace*{1.5em} - Readout subclass: cellular abundance stability proxy.\newline 
\hspace*{1.5em}Single-mutant context: \{\texttt{W108E}: -0.354, \texttt{G109P}: -1.062, \texttt{M34Q}: 0.605, \texttt{N35S}: 1.154, \texttt{...}\}.\newline 
\hspace*{1.5em}Candidate pool: \{\texttt{W108E+G109P}, \texttt{Y102P+M105K}, \texttt{M34Q+N35S}, \texttt{...}\}. \vspace{0.3em}\newline 
\textbf{Output:}\newline 
\hspace*{1.5em}Ranked candidate set: \texttt{M34Q+N35S}, \texttt{Y102P+M105K}, \texttt{W108E+G109P}, \texttt{...}. \\
\midrule
Statistics & Assays: 29; candidate number: 2,638; visible single-mutant context rows: 28,609. \\
\bottomrule
\end{tabular}
\caption{Task specification for \ntaskfour.}
\label{tab:task-t4}
\end{table*}

\begin{table}[t]
\centering
\renewcommand{\arraystretch}{1.15}
\setlength{\tabcolsep}{6pt}

\vspace{1em}

\begin{tabular}{p{0.28\linewidth}p{0.35\linewidth}cccc}
\toprule
\textbf{Objective} & \textbf{Metric} & \textbf{\ntone} &
\textbf{\nttwo} & \textbf{\ntthree} & \textbf{\ntfour} \\
\midrule
Global ranking
& Spearman Correlation
& $\times$ & $\checkmark$ & $\checkmark$ & $\checkmark$ \\

Top-weighted ranking
& NDCG
& $\times$ & $\checkmark$ & $\checkmark$ & $\checkmark$ \\

Peak discovery
& Normalized Maximum Score@$K$
& $\checkmark$ & $\checkmark$ & $\checkmark$ & $\checkmark$ \\

Peak coverage
& Recall@$K$
& $\checkmark$ & $\checkmark$ & $\checkmark$ & $\checkmark$ \\
\bottomrule
\end{tabular}
\caption{Evaluation objectives, corresponding metrics, and metric applicability
across the four tasks. Ranking metrics are not used for \ntone because it is a generation
task instead of a ranking task.}
\label{tab:evaluation_objectives}
\end{table}

\subsection{Evaluation Objectives}
\label{sec:task-objectives}

\Cref{tab:evaluation_objectives} summarizes the objective of each metric and its applicability to each task. Implementation details of the corresponding metrics are provided in \Cref{subsec:metrics}.

\begin{itemize}
\item \textbf{Global ranking with Spearman Correlation.}
Measures whether the predicted ordering agrees with the experimental ordering across the complete candidate pool, evaluating a model's ability to distinguish relative fitness throughout the library rather than only among the candidates placed at the top.

\item \textbf{Top-weighted ranking with NDCG.}
Measures the quality of the upper part of the predicted list by assigning greater importance to candidates near the top, evaluating whether a model places the most promising variants in positions that are most useful for downstream experimental testing.

\item \textbf{Peak discovery with Normalized Maximum Score@$K$.}
Measures the highest assay-normalized fitness among the $K$
submitted candidates, evaluating whether a model can identify at least one
strong candidate within the available budget of $K$ proposed mutations.

\item \textbf{Peak coverage with Recall@$K$.}
Measures how broadly a model recovers the ground-truth
top-performing candidates within a prediction budget $K$, complementing
peak discovery by assessing whether the model identifies multiple candidates
from the high-fitness region rather than only a single strong candidate.
\end{itemize}

\subsection{Data Construction}
\label{sec:data-construction}

\subsubsection{Assay Collection and Harmonization}
We assembled assay-level protein fitness landscapes from ProteinGym~\cite{notin2023proteingym}, MaveDB~\cite{esposito2019mavedb}, MegaScale~\cite{tsuboyama2023megascale}, FLAb~\cite{chungyoun2024flab}, Human Domainome~\cite{beltran2025domainome}, and CombinGym~\cite{chen2026combingym}. The collection also includes manually curated target DMS studies for CDKN2A~\cite{kimura2024cdkn2a}, SLC13A5~\cite{wang2025slc13a5}, and TrpB~\cite{johnston2024trpb}. These sources provide stability, binding, activity or organismal function measurements spanning single-mutant and combinatorial sequence--fitness landscapes. An assay is defined by its wild-type construct, experimental condition, readout, and score semantics, so measurements for one protein remain separate assays when they represent distinct experimental settings or engineering objectives. Each assay record contains biological and experimental metadata together with one or more assay-linked A3M alignments generated with MMseqs2~\cite{steinegger2017mmseqs2} against UniRef100~\cite{suzek2015uniref}. 
We curated an assay's primary task class, fitness type and readout subclass based on DMS databases and papers, where primary task class is artificially set to 3 different types: stability, binding, or activity function.

We harmonized mutant representations and score semantics across sources. A single substitution is encoded as \texttt{A12V}, and a multi-mutant as a plus-separated combination such as \texttt{A12V+G35L}. We retained mutants composed of the 20 standard amino acids and removed synonymous substitutions, stop codons, deletions, non-finite scores, repeated mutation sites within a mutant, and mutations that could not be reconstructed from the assay-specific wild-type sequence. Repeated measurements of the same canonical mutant within an assay were averaged. Assay-specific score semantics are retained for the score-construction step below.

For each assay, we calculate \texttt{DMS\_score} from the assay-level effect values obtained from the source file or reconstructed from the source readout specified in the assay metadata. Published processed effect columns are imported directly; assays requiring reconstruction are converted according to their measurement semantics. For example, affinity values reported as $K_{\mathrm{d}}$ are expressed as $-\log_{10}(K_{\mathrm{d}})$, and expression ratios are expressed as $\log_{10}(\mathrm{ER})$. The resulting effect values are standardized within each assay using a population z-score.
For readouts in which lower values indicate better performance, the corresponding effect values are sign-reversed, so larger \texttt{DMS\_score} values consistently indicate better assay performance. The resulting assay-level scores populate the measured single-mutant tables and the candidate records used by \ntone--\ntfour.

\subsubsection{Quality Control}

Every constructed assay--task instance underwent three groups of quality-control checks covering instance integrity, task consistency, and evaluation reliability:

\vspace{0.5em}
\begin{itemize}
\item \textbf{Instance integrity.} We first verified the correctness and completeness of each individual instance. Checks included mutation--sequence consistency, finite fitness values, complete metadata, absence of duplicate mutants or rows, and valid links to the corresponding A3M alignments. These checks ensure that each instance faithfully represents the underlying assay data, contains information required for downstream evaluation, and avoids malformed or incomplete records that could introduce artificial errors during model inference or metric computation throughout the evaluation pipeline.

\item \textbf{Task consistency.} We then validated whether each assay was correctly instantiated under the corresponding task definition. This included checking task-specific inputs, candidate-set construction, and mutation constraints, and task-specific formatting and consistency requirements. For assays shared across multiple task interfaces, we additionally cross-checked sequence, measurement, and visual data to ensure that the same underlying assay information remained consistent across tasks and that no discrepancies were introduced during task-specific preprocessing or instance conversion.

\item \textbf{Evaluation reliability.} Finally, we examined whether individual instances supported stable and meaningful assessment across all reported evaluation metrics. We excluded severe tie-heavy \nttwo--\ntfour instances and \ntone instances in which exact top-region ties made top-$K$ membership ambiguous. Local near-ties that did not alter the relevant ranking boundary were retained and documented. We also tested for accidental order leakage by measuring monotonicity and rank correlation between row position and \dmsscore, ensuring that benchmark ordering did not provide unintended information about the target labels or create spurious performance gains unrelated to genuine mutant prioritization ability.
\end{itemize}
\vspace{0.5em}

After quality filtering, \textbf{\bench} comprises 202 unique assays, 293 assay--task instances, and 607,269 target candidate rows. Per-task instance counts are listed in the task specification tables. Since tasks impose different evidence and candidate constraints, the same assay may contribute to multiple tasks.
At the assay--task level, the current benchmark contains 124 stability, 101 activity or organismal function, and 68 binding instances. Its source composition comprises 76 MegaScale, 78 ProteinGym, 44 MaveDB, 34 FLAb, 36 Human Domainome, 19 CombinGym and 6 manually curated DMS instances. 

%% file: sections/04experiments.tex
\clearpage

\section{Experimental Setup}
\label{sec:experimental_setup}

\subsection{Baselines}

We compare four baseline families: random selection, zero-shot protein language models, standalone general-purpose LLMs, and LLM-based scientific agents.

\vspace{-0.3em}
\paragraph{Random}
For \ntone, the random baseline uniformly samples $K$ legal single substitutions without replacement. For \nttwo--\ntfour, it produces a uniformly random permutation of the supplied candidate pool.

\vspace{-0.3em}
\paragraph{Protein Language Models}
We evaluated six zero-shot protein language models spanning complementary biological input modalities. ESM-2 (650M)~\cite{lin2023evolutionary} and ProGen2-base (764M)~\cite{nijkamp2023progen2} use protein sequences alone; ProSST-2048~\cite{li2024prosst} incorporates discrete structure tokens; S3F~\cite{zhang2024s3f} integrates sequence, backbone, and surface representations; and VenusREM~\cite{tan2025venusrem} additionally incorporates MSA-derived evolutionary information. S3F-MSA combines S3F with an ensemble of five independently trained EVE models~\cite{frazer2021eve}.

\vspace{-0.3em}
\paragraph{LLMs}
We evaluated six general-purpose LLMs from diverse model families: GPT-6 Astra~\cite{openai2026gpt6astra}, Claude Opus 5~\cite{anthropic2026claude5}, Gemini 3.1 Pro~\cite{google2026gemini3}, Kimi K3~\cite{kimiteam2026kimik3openfrontier}, GLM-5.2~\cite{zai2026glm52}, and DeepSeek-V4-Pro~\cite{deepseek2026v4}. All models were evaluated without any fine-tuning. For each task, they received the same prompt template and the same model-visible inputs defined in \Cref{sec:task-suite}, and were required to return predictions in the prescribed structured format.

\vspace{-0.3em}
\paragraph{Agents}
We additionally evaluated scientific agent frameworks Biomni~\cite{huang2025biomni} and our \textbf{\toolkit} toolkit. They received the same task-visible inputs and followed the same structured-output requirements as the standalone LLMs, while retaining the ability to orchestrate the tools and scientific resources provided by their frameworks.

\subsection{Metrics}
\label{subsec:metrics}

We compute all metrics independently for each assay-level instance and then macro-average the resulting values, so that every instance contributes equally to the reported performance. After score harmonization during data construction, larger \dmsscore values consistently indicate better fitness. We use task-specific values of $K$ to account for differences in task formulation and candidate-space size. In \ntone, models generate candidates from a large single-mutant space, so we set $K=40$ to provide a sufficiently broad generation budget. In \nttwo--\ntfour, models rank a smaller supplied candidate pool, so we set $K=5$ to maintain a selective evaluation of the top-ranked candidates within the reduced candidate space.

\vspace{-0.3em}
\paragraph{Spearman Correlation}
Measures agreement between the predicted and experimental rankings over the complete candidate pool. For an instance with $n$ candidates, we assign each candidate an experimental rank and a predicted rank, and compute the Pearson correlation between the two rank vectors. Ties in the experimental scores are assigned their average rank.

\vspace{-0.3em}
\paragraph{NDCG}
Evaluates ranking quality by placing greater weight on high-fitness candidates near the top of the list. We first apply min-max normalization to the \dmsscore values to map them into relevance scores $r_i = (s_i - s_{\min}) / (s_{\max} - s_{\min}) \in [0, 1]$. For a predicted permutation $\pi$ of $n$ candidates, NDCG is defined as:
\[
\operatorname{NDCG}(\pi)=\frac{\operatorname{DCG}(\pi)}{\operatorname{IDCG}},
\qquad
\operatorname{DCG}(\pi)=\sum_{j=1}^{n}\frac{2^{r_{\pi(j)}}-1}{\log_2(j+1)},
\]
and $\operatorname{IDCG}$ represents the ideal maximum possible $\operatorname{DCG}$, computed by evaluating the same discounted gain formula after sorting all candidates in decreasing order of true relevance $r_i$.

\vspace{-0.3em}
\paragraph{Normalized Maximum Score} Measures peak discovery among the submitted candidates. Let $\mathcal{H}_K$ denote the legal candidates occurring within the first $K$ submitted positions. We define
\[
\operatorname{NMS}@K
=
\frac{\max_{i\in\mathcal{H}_K}s_i-s_{\min}}
{s_{\max}-s_{\min}}.
\]
For \ntone, $\mathcal{H}_{40}$ is the submitted single-mutant set under the
fixed generation budget. In this task, $s_{\min}$ and $s_{\max}$ denote the
minimum and maximum experimental scores over the full measured single-mutant
space. For \nttwo--\ntfour, $\mathcal{H}_{5}$ consists of the leading
candidates in the predicted multi-mutant ranking, and $s_{\min}$ and
$s_{\max}$ are computed over the corresponding supplied candidate pool.

\vspace{-0.3em}
\paragraph{Recall}
Measures coverage of the experimentally high-fitness region. Let $\mathcal{T}_K$ denote the ground-truth top-$K$ candidates, with all candidates tied at the ground-truth cutoff included as relevant. For an instance with candidate-pool size $n$, we compute Recall@$K$ as:

\vspace{-0.3em}
\[
\operatorname{Recall}@K
=
\frac{|\mathcal{H}_K\cap\mathcal{T}_K|}
{\min(K,n)}.
\]
\vspace{-0.3em}

The denominator remains $\min(K,n)$ when ties expand the ground-truth relevant set. Recall complements NMS, which depends only on the strongest recovered candidate.

LLM and agent outputs are not fully controllable and may include refusals, incomplete responses, or outputs that cannot be parsed successfully. The post-processing procedures for missing or incomplete assay responses are described in~\Cref{app:llm_missing_assays} and applied uniformly across all models.

\subsection{Implementation Details}
\label{sec:implementation-details}

We used a common evaluation pipeline across all baseline families. Task-visible inputs and required output formats followed \Cref{sec:task-suite}, while the model-specific inference procedures are described below.

\paragraph{PLM Inference}
PLM inference used label-free, zero-shot mutation scores under fixed model-specific configurations, without natural-language assay descriptions or task-specific auxiliary information. Wild-type structures were generated with AlphaFold~3~\cite{abramson2024alphafold3}, and evolutionary alignments were constructed against UniRef100~\cite{suzek2015uniref} using MMseqs2~\cite{steinegger2017mmseqs2}. For \ntone, the scores were used to rank all legal single substitutions; for \nttwo--\ntfour, they were used to rank the supplied candidate sets. Model-specific checkpoints, scoring definitions, and protocols for multi-substitution, multi-chain, and long-sequence inputs are documented in \Cref{app:protein-model-deployment}.

\paragraph{LLM Inference}
Each prompt was instantiated from a task-specific template, provided in \Cref{app:prompt_templates}, and submitted as a single non-streaming request through an asynchronous OpenAI-compatible client. 
The models \texttt{gpt-6-astra}, \texttt{claude-opus-5}, \texttt{gemini-3.1-pro-preview}, \texttt{kimi-k3}, \texttt{glm-5.2}, and \texttt{deepseek-v4-pro} were required to return a task-specific structured output: a proposed mutation set for \ntone and a proposed ranking for \nttwo--\ntfour.
Default maximum completion lengths were set to 65,536 tokens for \texttt{gemini-3.1-pro-preview}, \texttt{kimi-k3} and \texttt{glm-5.2}, and 32,768 tokens for the remaining models.
Sampling parameters and reasoning-effort settings were left at their API defaults for all evaluated model configurations.
Responses that could be successfully parsed into the required structure were recorded as results, without repairing, replacing, or filtering predicted mutations before evaluation. 
Failed instances were retried for up to 10 complete inference runs, until all pending instances succeeded or a run produced no additional successful results.

\paragraph{Agent Inference}
We used Biomni~\cite{huang2025biomni} to instantiate each instance as an independent task with a task-specific agent prompt. We evaluated two backbone models: \texttt{gpt-6-astra} and \texttt{claude-opus-5}. Unlike standalone LLM inference, each agent operated in a multi-turn loop with iterative tool use and intermediate feedback. At each turn, the agent either emitted an \texttt{\textless execute\textgreater} block to invoke a Biomni tool, inspect public database or model outputs, or run focused Python/Bash code, or emitted a final \texttt{\textless solution\textgreater} block. After each tool-use turn, the observation was returned to the agent before the next turn. We required at least one execution--observation round before accepting a final answer and allowed at most 40 tool-use rounds. Each model call was limited to a maximum of 24,576 output tokens. Once the tool-use budget was exhausted, further tool calls were disabled and the agent was required to return the final answer in the specified structured-JSON format. GPU-intensive Biomni tools and eligible code-execution blocks were dispatched to a remote GPU worker, while the agent remained responsible for selecting actions and producing the final ranking. Only the structured JSON object within the \texttt{\textless solution\textgreater} block was parsed as the final response.

\begin{figure*}[t]
  \centering
  \includegraphics[width=\textwidth]{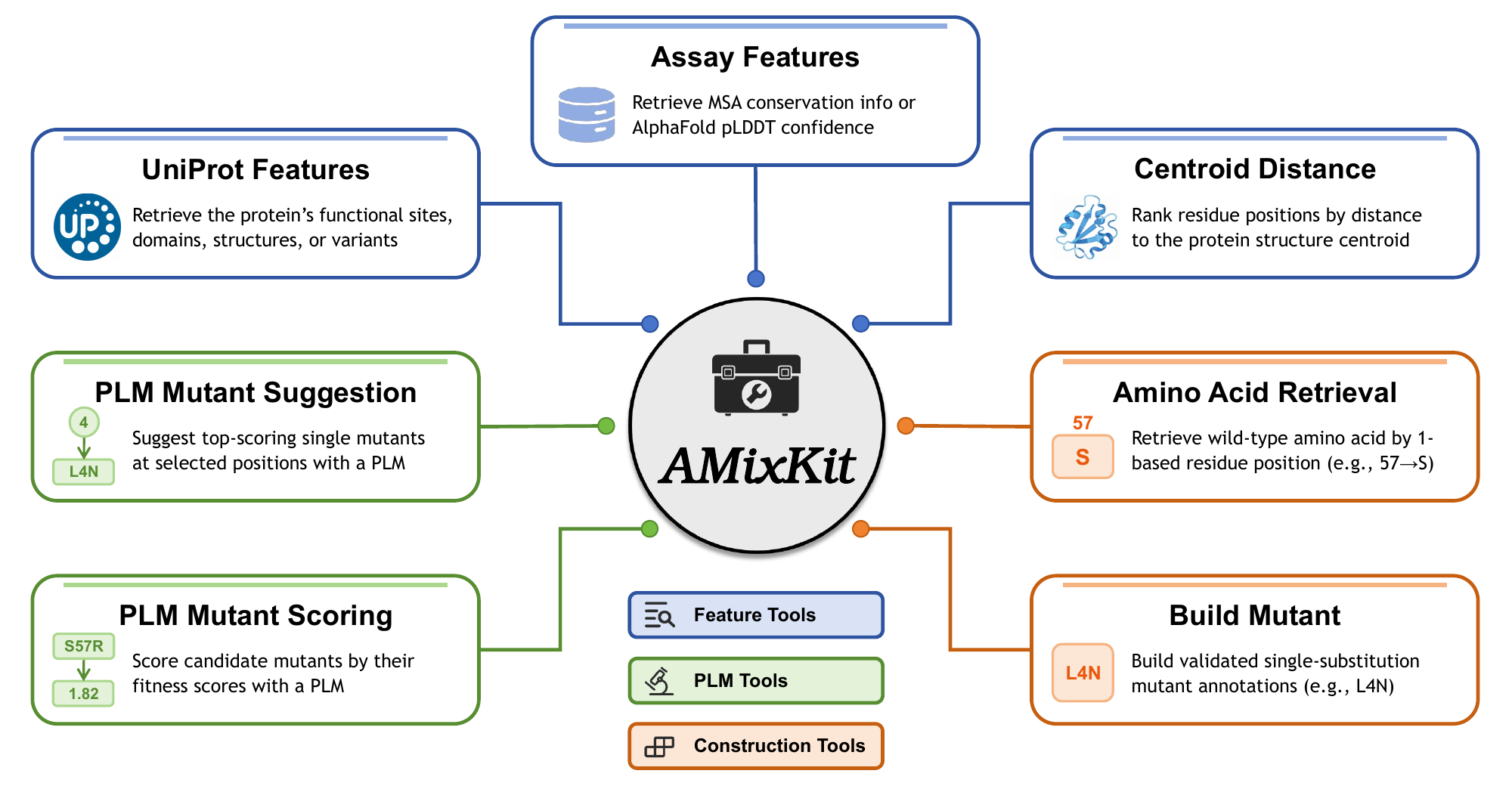}
  \caption{\textbf{Overview of \toolkit, integrating protein features, PLM inference, and mutant construction.}
  }
  \label{fig:amixkit}
\end{figure*}

\paragraph{AMixKit Toolset}
We additionally evaluated agents equipped with \textbf{\toolkit}, a lightweight toolkit that provides standardized protein-specific operations for evidence retrieval, mutation construction, and PLM-based prediction. As illustrated in \Cref{fig:amixkit}, \textbf{\toolkit} contains seven tools:

\begin{itemize}
    \item \textbf{UniProt Features:} retrieves functional sites, domains, structural annotations, natural variants, and curated mutagenesis records for a specified protein accession.

    \item \textbf{Assay Features:} summarizes assay-specific evolutionary or structural evidence, including MSA coverage and conservation or AlphaFold pLDDT confidence for an exact benchmark assay.

    \item \textbf{Centroid Distance:} ranks residue positions by their distance to the structural centroid, with chain-aware handling of multichain proteins, to identify structurally central or peripheral sites.

    \item \textbf{Amino Acid Retrieval:} retrieves the wild-type amino acid at specified sequence positions.

    \item \textbf{Build Mutant:} constructs and validates single-substitution annotations from selected positions and target amino acids while enforcing wild-type sequence consistency.

    \item \textbf{PLM Mutant Suggestion:} generates and ranks the 19 possible non-wild-type substitutions at selected positions using a specified PLM and reports the highest-scoring candidates per position.

    \item \textbf{PLM Mutant Scoring:} assigns PLM-based fitness scores to a set of single- or multi-mutant candidates.
\end{itemize}

The PLM tools support ESM-2, ProGen2, ProSST, S3F, S3F-MSA, and VenusREM, using cached scores when available and the scoring service otherwise. Collectively, these tools allow agents to gather biological evidence, construct valid mutations, and incorporate protein-model predictions in one workflow. \textbf{\toolkit} agents followed the same multi-turn execution--observation protocol as the Biomni agents.

We evaluate \textbf{\toolkit} with both frontier LLMs and our in-house model, AMix-2.1. Frontier LLMs demonstrate strong capabilities, but may still exhibit over-refusal or false refusals on biology-related tasks. Moreover, their API costs can become substantial when deployed at scale. Motivated by these limitations, we developed AMix-2.1 by scaling AMix-2~\cite{qiu2026amix} to hundreds of billions of parameters and applying multi-task agentic RL. We will release further technical details and open-source both AMix-2.1 and \textbf{\toolkit} in future work.

\clearpage

\section{Main Results}
\label{sec:main}

\begin{figure*}[t]
  \centering
  \includegraphics[width=\textwidth]{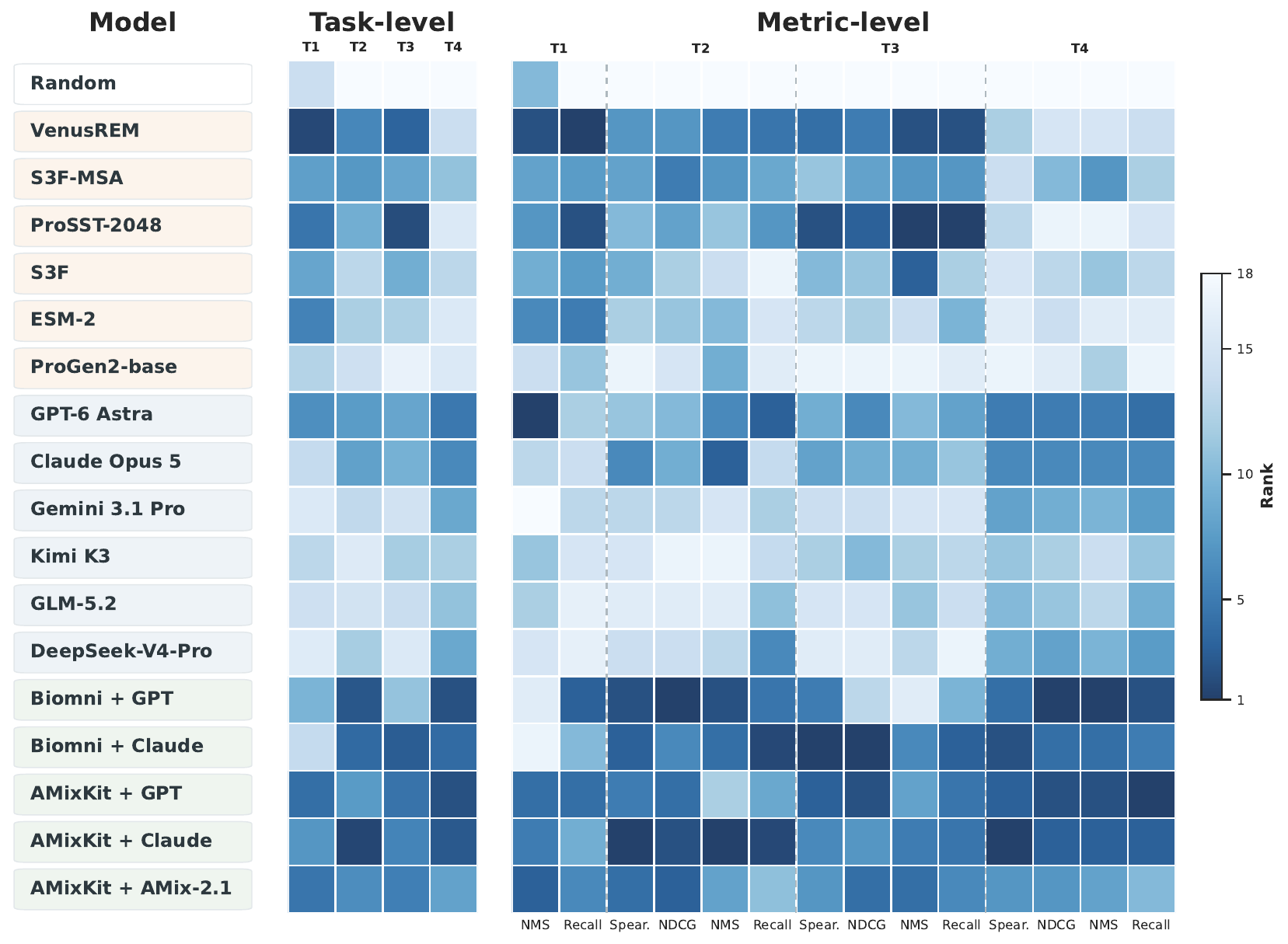}
  
  \caption{\textbf{Task-level and metric-level rank-based model capability.} For each task and metric, all methods are ranked according to their performance, with $rank=1$ indicating the best method.}
  \label{fig:overall_results}
  \end{figure*}

\subsection{Overall Results}
\label{sec:overall_analysis}
We use \Cref{fig:overall_results} to summarize benchmark performance at two levels: at the \textbf{metric} level, each method is ranked based on each metric for fine-grained comparison; at the \textbf{task} level, metric-specific ranks are aggregated into a task-level mean rank for comprehensive performance. 
The random baseline, PLMs, LLMs, and tool-augmented agents are compared rank-based, demonstrating how method performance changes with task formulation, available experimental evidence, and evaluation objective. 
Two qualitative observations emerge:


\begin{itemize}
    \item \textbf{At the task level, relative strengths vary with task formulation and available evidence.}
    In \ntone, PLMs lead overall. Agents lead in \nttwo, whereas \ntthree shows a more interleaved ordering between PLMs and agents. In \ntfour, agents achieve their clearest overall advantage under comprehensive target-specific evidence. This non-monotonic pattern highlights task formulation alongside experimental evidence availability.
    \item \textbf{At the metric level, overall standing does not guarantee leadership on every metric.}
    In \ntone, PLMs hold the strongest overall positions, yet GPT-6 Astra leads NMS@$40$.
    In \ntthree, agents rank highest on global and top-weighted ordering, whereas ProSST-2048 ranks highest on peak discovery and top-candidate recovery. Even where agents lead at the task level, as in \nttwo and \ntfour, the leading agent configuration varies across the four evaluation metrics. Task-level mean ranks therefore summarize overall strength but cannot identify the best method for every evaluation objective.

\end{itemize}

\begin{table}[t]
\centering
\small
\renewcommand{\arraystretch}{1.05}

\begin{tabular}{l >{\rmfamily}l
    S[table-format=1.4]
    S[table-format=1.4]
}

\toprule
\textbf{Category} & \textbf{Model} & \textbf{NMS@$40$} & \textbf{Recall@$40$} \\
\midrule

\addlinespace[0.2em]
\multirow{1}{*}{\textbf{Statistic}}
  & Random & 0.7749 & 0.0142 \\
\addlinespace[0.2em]

\midrule

\addlinespace[0.2em]
\multirow{6}{*}{\textbf{PLM}}
& VenusREM & \underline{0.7932} & \textbf{0.0427} \\
& S3F-MSA & 0.7816 & 0.0364 \\
& ProSST-2048 & 0.7847 & \underline{0.0407} \\
& S3F & 0.7804 & 0.0364 \\
& ESM-2  & 0.7854 & 0.0370 \\
& ProGen2-base & 0.7709 & 0.0274 \\
\addlinespace[0.2em]
\midrule

\addlinespace[0.2em]
\multirow{6}{*}{\textbf{LLM}}
  & GPT-6 Astra & \textbf{0.7966} & 0.0254 \\
  & Claude Opus 5 & 0.7729 & 0.0205 \\
  & Gemini 3.1 Pro & 0.7620 & 0.0211 \\
  & Kimi K3 & 0.7739 & 0.0199 \\
  & GLM-5.2 & 0.7734 & 0.0185 \\
  & DeepSeek-V4-Pro & 0.7703 & 0.0185 \\
\addlinespace[0.2em]

\midrule

\addlinespace[0.2em]
\multirow{5}{*}{\textbf{Agent}}
& Biomni $+$ GPT-6 Astra & 0.7677 & 0.0388 \\
& Biomni $+$ Claude Opus 5 & 0.7651 & 0.0307 \\
\addlinespace[0.2em]
& \toolkit $+$ GPT-6 Astra     & 0.7866 & 0.0386 \\
& \toolkit $+$ Claude Opus 5   & 0.7858 & 0.0354 \\
& \toolkit $+$ AMix-2.1        & 0.7872 & 0.0368 \\
\addlinespace[0.2em]

\bottomrule
\end{tabular}
\caption{Performance on \ntaskone. The best results are in \textbf{bold} and the second-best results are \underline{underlined}. The same convention is used in the tables below.}
\label{tab:main_results_t1}
\end{table}

\subsection{Task-specific Results}
\label{sec:task_specific_analysis}


\subsubsection{T1: Single-mutant generation}
As shown in \Cref{tab:main_results_t1}, single-mutant generation remains challenging under a fixed generation budget, with limited recovery of top-ranked mutants and only modest gains in peak discovery over random selection.
Learned methods achieve higher Recall@$40$ than random selection, but the best value increases from only $0.0142$ to $0.0427$, still recovering fewer than 5\% of the ground-truth top-40 mutations. 
The relative gains in NMS@$40$ are noticeably smaller, rising at most from $0.7749$ to $0.7966$, with numerous models showing no competitive edge over random guessing in this setting.

Across model categories, all six PLMs achieve higher Recall@$40$ than the evaluated standalone LLMs. PLMs also outperform standalone LLMs in NMS@$40$, with the exception of ProGen2-base and GPT-6 Astra, while \textbf{\toolkit} agents outperform all PLMs except VenusREM. Taken together, PLMs remain competitive in measurement-free mutant generation, while frontier LLMs and \textbf{\toolkit} tool-use supports identification of highly promising individual candidates.

\begin{table}[p]
\centering
\small
\renewcommand{\arraystretch}{1.05}

\begin{tabular}{l >{\rmfamily}l
    S[table-format=1.4]
    S[table-format=1.4]
    S[table-format=1.4]
    S[table-format=1.4]
}

\toprule
\textbf{Category} & \textbf{Model} & \textbf{Spearman} & \textbf{NDCG} & \textbf{NMS@$5$} & \textbf{Recall@$5$} \\
\midrule

\addlinespace[0.2em]
\multirow{1}{*}{\textbf{Statistic}}
  & Random & -0.0215 & 0.8196 & 0.7113 & 0.0811 \\
\addlinespace[0.2em]

\midrule

\addlinespace[0.2em]
\multirow{6}{*}{\textbf{PLM}}
& VenusREM & 0.2804 & 0.8684 & 0.8003 & 0.2162 \\
& S3F-MSA & 0.2803 & 0.8700 & 0.7964 & 0.2027 \\
& ProSST-2048 & 0.2777 & 0.8657 & 0.7829 & 0.2108 \\
& S3F & 0.2785 & 0.8626 & 0.7751 & 0.1757 \\
& ESM-2  & 0.2539 & 0.8634 & 0.7831 & 0.1892 \\
& ProGen2-base & 0.1405 & 0.8563 & 0.7887 & 0.1784 \\
\addlinespace[0.2em]

\midrule

\addlinespace[0.2em]
\multirow{6}{*}{\textbf{LLM}}
  & GPT-6 Astra & 0.2726 & 0.8635 & 0.7967 & \underline{0.2216} \\
  & Claude Opus 5 & 0.2841 & 0.8653 & 0.8031 & 0.1919 \\
  & Gemini 3.1 Pro & 0.2239 & 0.8595 & 0.7723 & 0.1973 \\
  & Kimi K3 & 0.1973 & 0.8507 & 0.7499 & 0.1919 \\
  & GLM-5.2 & 0.1607 & 0.8519 & 0.7706 & 0.2000 \\
  & DeepSeek-V4-Pro & 0.2187 & 0.8587 & 0.7760 & 0.2135 \\
\addlinespace[0.2em]

\midrule

\addlinespace[0.2em]
\multirow{5}{*}{\textbf{Agent}}
& Biomni $+$ GPT-6 Astra & \underline{0.3194} & \textbf{0.8767} & \underline{0.8038} & 0.2162 \\
& Biomni $+$ Claude Opus 5 & 0.3064 & 0.8689 & 0.8016 & \textbf{0.2351} \\
\addlinespace[0.2em]
& \toolkit $+$ GPT-6 Astra     & 0.2851 & 0.8702 & 0.7781 & 0.2027 \\
& \toolkit $+$ Claude Opus 5   & \textbf{0.3330} & \underline{0.8764} & \textbf{0.8120} & \textbf{0.2351} \\
& \toolkit $+$ AMix-2.1        & 0.2960 & 0.8706 & 0.7911 & 0.2000 \\
\addlinespace[0.2em]

\bottomrule
\end{tabular}
\caption{Performance on \ntasktwo.}
\label{tab:main_results_t2}

\vspace{3em}

\begin{tabular}{l >{\rmfamily}l
    S[table-format=1.4]
    S[table-format=1.4]
    S[table-format=1.4]
    S[table-format=1.4]
}

\toprule
\textbf{Category} & \textbf{Model} & \textbf{Spearman} & \textbf{NDCG} & \textbf{NMS@$5$} & \textbf{Recall@$5$} \\
\midrule

\addlinespace[0.2em]
\multirow{1}{*}{\textbf{Statistic}}
  & Random & 0.0411 & 0.8216 & 0.7659 & 0.1970 \\
\addlinespace[0.2em]

\midrule

\addlinespace[0.2em]
\multirow{6}{*}{\textbf{PLM}}
& VenusREM & 0.3698 & 0.8897 & \underline{0.8578} & \underline{0.3612} \\
& S3F-MSA & 0.3159 & 0.8860 & 0.8382 & 0.3224 \\
& ProSST-2048 & \underline{0.3836} & 0.8903 & \textbf{0.8620} & \textbf{0.3672} \\
& S3F & 0.3357 & 0.8842 & 0.8523 & 0.2955 \\
& ESM-2  & 0.3128 & 0.8840 & 0.8099 & 0.3045 \\
& ProGen2-base & 0.1870 & 0.8606 & 0.7835 & 0.2716 \\
\addlinespace[0.2em]

\midrule

\addlinespace[0.2em]
\multirow{6}{*}{\textbf{LLM}}
  & GPT-6 Astra & 0.3516 & 0.8886 & 0.8258 & 0.3134 \\
  & Claude Opus 5 & 0.3540 & 0.8858 & 0.8263 & 0.3015 \\
  & Gemini 3.1 Pro & 0.2865 & 0.8814 & 0.8019 & 0.2806 \\
  & Kimi K3 & 0.3131 & 0.8857 & 0.8124 & 0.2925 \\
  & GLM-5.2 & 0.2676 & 0.8701 & 0.8149 & 0.2866 \\
  & DeepSeek-V4-Pro & 0.2627 & 0.8698 & 0.8106 & 0.2687 \\
\addlinespace[0.2em]

\midrule

\addlinespace[0.2em]
\multirow{5}{*}{\textbf{Agent}}
& Biomni $+$ GPT-6 Astra & 0.3687 & 0.8828 & 0.7925 & 0.3045 \\
& Biomni $+$ Claude Opus 5 & \textbf{0.4511} & \textbf{0.8985} & 0.8511 & 0.3463 \\
\addlinespace[0.2em]
& \toolkit $+$ GPT-6 Astra     & 0.3833 & \underline{0.8907} & 0.8323 & 0.3433 \\
& \toolkit $+$ Claude Opus 5   & 0.3665 & 0.8875 & 0.8515 & 0.3433 \\
& \toolkit $+$ AMix-2.1        & 0.3616 & 0.8902 & 0.8515 & 0.3254  \\
\addlinespace[0.2em]

\bottomrule
\end{tabular}
\caption{Performance on \ntaskthree.}
\label{tab:main_results_t3}
\end{table}

\subsubsection{T2: Measurement-free multi-mutant ranking} As shown in \Cref{tab:main_results_t2}, all learned measurement-free ranking methods outperform the random baseline across all metrics, with the best Spearman correlation improving from $-0.0215$ to $0.3330$. However, the best Recall@$5$ reaches only $0.2351$, recovering fewer than one quarter of the ground-truth top-five candidates. While models can extract useful information regarding multi-mutant fitness, they cannot reliably reconstruct the ordering or consistently recover the most promising candidates without target-specific measurements.

Agents achieve the best score on all four \nttwo metrics. \toolkit $+$ Claude Opus 5 achieves the highest Spearman correlation, NMS@$5$, and Recall@$5$, at $0.3330$, $0.8120$, and $0.2351$, while Biomni $+$ GPT-6 Astra achieves the highest NDCG at $0.8767$. The strongest non-agent competitors are distributed across the LLM and PLM families: Claude Opus 5 is competitive in Spearman and NMS@$5$, whereas S3F-MSA leads in NDCG.

\subsubsection{T3: Anchor-informed multi-mutant ranking} 
As shown in \Cref{tab:main_results_t3}, models of all categories perform better under the anchor-informed setting, as compared to the measurement-free setting. All evaluated methods outperform random ranking across the four metrics, showing that anchor-conditioned reasoning can improve multi-mutant prioritization, while the effects of additional mutations remain difficult to infer reliably.

The leading method depends on the evaluation objective, revealing complementary strengths of agents and PLMs. Biomni $+$ Claude Opus 5 achieves the highest Spearman correlation at $0.4511$ and NDCG at $0.8985$; whereas ProSST-2048 attains the highest NMS@$5$ and Recall@$5$, with scores of $0.8620$ and $0.3672$, respectively, and outperforms most other agents across metrics. 
When an anchor measurement is available, no model family consistently dominates; therefore, the choice of model should be guided by specific experiment objectives.

\begin{table}[t]
\centering
\small
\renewcommand{\arraystretch}{1.05}

\begin{tabular}{l >{\rmfamily}l
    S[table-format=1.4]
    S[table-format=1.4]
    S[table-format=1.4]
    S[table-format=1.4]
}

\toprule
\textbf{Category} & \textbf{Model} & \textbf{Spearman} & \textbf{NDCG} & \textbf{NMS@$5$} & \textbf{Recall@$5$} \\
\midrule

\addlinespace[0.2em]
\multirow{1}{*}{\textbf{Statistic}}
  & Random & -0.0204 & 0.8124 & 0.6371 & 0.0552 \\
\addlinespace[0.2em]

\midrule

\addlinespace[0.2em]
\multirow{6}{*}{\textbf{PLM}}
& VenusREM & 0.3821 & 0.8809 & 0.8212 & 0.2552 \\
& S3F-MSA & 0.3662 & 0.9005 & 0.8686 & 0.2897 \\
& ProSST-2048 & 0.3779 & 0.8725 & 0.8078 & 0.2483 \\
& S3F & 0.3487 & 0.8836 & 0.8410 & 0.2759 \\
& ESM-2  & 0.3153 & 0.8828 & 0.8129 & 0.2276 \\
& ProGen2-base & 0.2568 & 0.8736 & 0.8357 & 0.1793 \\
\addlinespace[0.2em]

\midrule

\addlinespace[0.2em]
\multirow{6}{*}{\textbf{LLM}}
  & GPT-6 Astra & 0.5988 & 0.9274 & 0.8935 & 0.4483 \\
  & Claude Opus 5 & 0.5823 & 0.9244 & 0.8723 & 0.4276 \\
  & Gemini 3.1 Pro & 0.5329 & 0.9059 & 0.8595 & 0.4000 \\
  & Kimi K3 & 0.4794 & 0.8956 & 0.8337 & 0.3586 \\
  & GLM-5.2 & 0.4959 & 0.8984 & 0.8338 & 0.3862 \\
  & DeepSeek-V4-Pro & 0.5308 & 0.9059 & 0.8595 & 0.4000 \\
\addlinespace[0.2em]

\midrule

\addlinespace[0.2em]
\multirow{5}{*}{\textbf{Agent}}
& Biomni $+$ GPT-6 Astra & 0.6373 & \textbf{0.9347} & \textbf{0.9145} & \underline{0.4828} \\
& Biomni $+$ Claude Opus 5 & \underline{0.6446} & 0.9293 & 0.9015 & 0.4345 \\
\addlinespace[0.2em]
& \toolkit $+$ GPT-6 Astra     & 0.6382 & \underline{0.9336} & \underline{0.9131} & \textbf{0.4897} \\
& \toolkit $+$ Claude Opus 5   & \textbf{0.6462} & \underline{0.9336} & 0.9090 & 0.4690 \\
& \toolkit $+$ AMix-2.1        & 0.5531 & 0.9115 & 0.8642 & 0.3862  \\
\addlinespace[0.2em]

\bottomrule
\end{tabular}
\caption{Performance on \ntaskfour.}
\vspace{-0.1in}
\label{tab:main_results_t4}
\end{table}

\subsubsection{T4: Single-mutant-informed multi-mutant ranking} 
As shown in \Cref{tab:main_results_t4}, providing candidate-complete single-mutant fitness context shifts the strongest results toward LLM-based systems, particularly tool-augmented agents, which generally outperform PLMs across all metrics. Unlike \ntthree, which provides the measured score of one anchor variant, this task supplies evidence for every candidate component, directly testing whether models can integrate distributed assay information. Nevertheless, single-mutant scores do not uniquely determine multi-mutant fitness because non-additive interactions may alter substitution effects in combined sequences. Thus, this setting highlights both the value of component-level experimental evidence and the challenge of extrapolating it to combinatorial variants.

Tool augmentation through both Biomni and \textbf{\toolkit} consistently improves performance across metrics and backbones, while the best agent depends on the objective.
\toolkit $+$ Claude Opus 5 achieves the highest Spearman correlation at $0.6462$; Biomni $+$ GPT-6 Astra leads in NDCG at $0.9347$ and NMS@$5$ at $0.9145$; and \toolkit $+$ GPT-6 Astra attains the highest Recall@$5$ at $0.4897$.
Tool-augmented agents demonstrate the broadest advantage on \ntfour, while the differing leaders for global ordering, top-weighted ranking, peak discovery, and candidate coverage underscore that these metrics capture complementary ranking aspects.

%% file: sections/05analysis.tex
\section{Analysis}
\label{sec:analysis}

\subsection{Comparing PLMs and LLMs}
\label{sec:science}

As discussed in \Cref{sec:main}, PLMs and LLMs exhibit distinct performance trajectories across task environments. Mechanistically, this divergence stems from their contrasting modes of information processing: PLMs rely primarily on sequential and homologous signals, whereas LLMs demonstrate superior capacity in leveraging contextual and textual auxiliary conditions. From a biological perspective, the underlying driver of this disparity is task-dependent, displaying a pronounced demarcation between the single-mutant task of \ntone and multi-mutant tasks of \nttwo--\ntfour. Nevertheless, both model families encounter shared operational bottlenecks. Specifically, \Cref{subsec:t1-single-mutant-biology} investigates family-specific preferences for amino-acid mutation types; \Cref{subsec:multi-mutant-differences} reveals that the utility of single-mutant fitness values in LLMs and agentic systems is largely constrained to additive landscapes; and \Cref{subsec:shared-difficulties} evaluates how search space complexity governs performance across architectures. We systematically analyze each aspect in the subsequent sections.

\subsubsection{Divergent Single-mutant Proposal Strategies}
\label{subsec:t1-single-mutant-biology}

\begin{figure}[t]
  \centering
  \includegraphics[width=\textwidth]{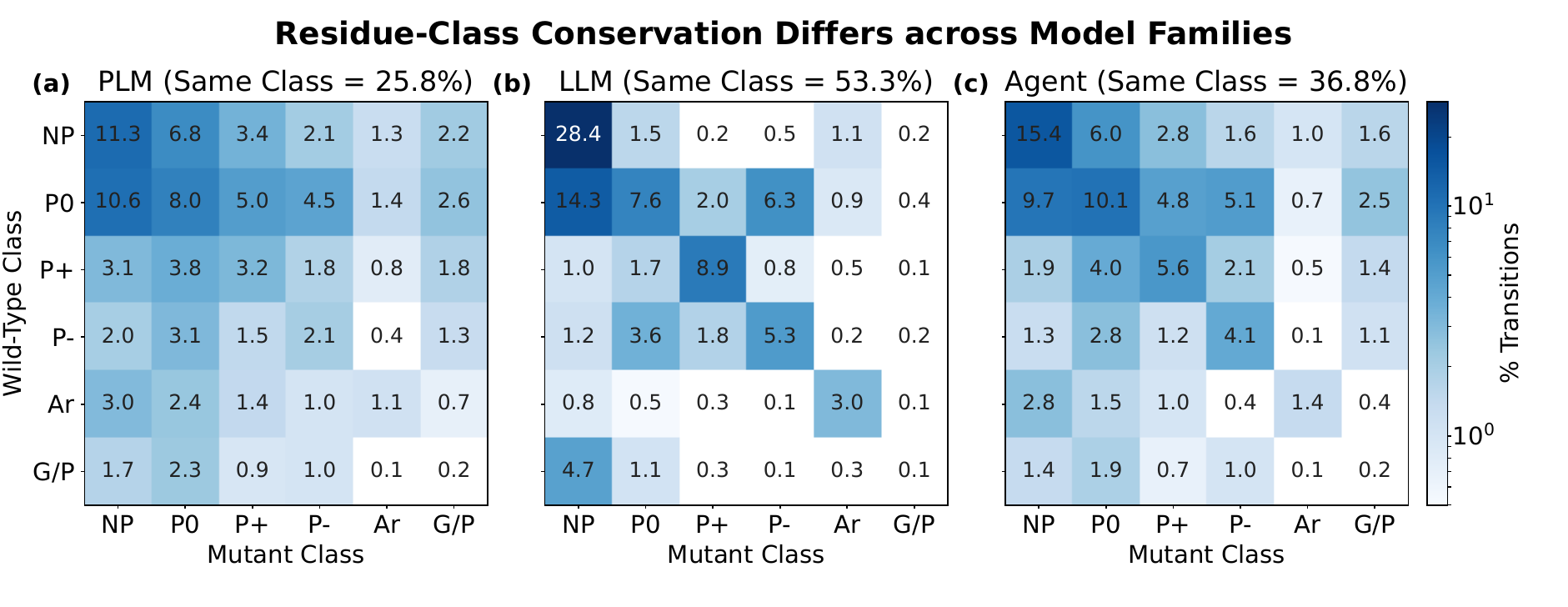}
  \caption{\textbf{Directional six-class amino-acid transitions across model families.} The three panels present $6\times6$ wild-type-to-mutant class transition matrices for PLMs, LLMs, and agents. The residue classes comprise nonpolar (NP), polar uncharged (P0), polar positively charged (P+), polar negatively charged (P-), aromatic (Ar), and Gly/Pro (G/P). Each cell denotes the percentage of successful \ntone hits within the respective model-family matrix (normalized to $100\%$). Color intensity follows an exponential scale, with frequencies below $0.5\%$ depicted in white. Charge-separated states are preserved, maintaining distinct off-diagonal transitions for acidic-to-basic and basic-to-acidic substitutions.}
  \label{fig:t1-aa-preferences}
\end{figure}

As \ntaskone lacks target-specific mutation measurements, models prioritize candidates using the supplied assay context and protein-specific evolutionary information when available; agents may also incorporate tool-derived evidence. We evaluated each model by the proportion of top-$30\%$ experimental hits within its top-$40$ predictions, aggregating results by model family. Mean success rates reached $48.0\%$ for PLMs, $42.5\%$ for LLMs, and $50.4\%$ for agents. While overall predictive accuracy remains comparable across architectures, the underlying mutational strategies diverge substantially.

To profile candidate selections, we quantified directed transitions across six physicochemical residue classes among successful variants: \textbf{nonpolar (NP)}, \textbf{polar uncharged (P0)}, \textbf{polar positively charged (P+)}, \textbf{polar negatively charged (P-)}, \textbf{aromatic (Ar)}, and \textbf{Gly/Pro (G/P)}. As shown in \Cref{fig:t1-aa-preferences}, the $6\times6$ transition matrices capture all 36 wild-type-to-mutant class shifts, normalized within each model family. Distinguishing P+ from P- preserves critical charge-reversal dynamics, whereas Gly and Pro are grouped together for compactness.

Class-preserving substitutions along the main diagonal accounted for $25.8\%$ of successful PLM proposals, compared to $53.3\%$ for LLMs and $36.8\%$ for agents. Successful LLM proposals were thus predominantly conservative, whereas PLM successes spanned a broader spectrum of class-altering transitions, with LLM-based agents occupying an intermediate regime. This disparity is not an artifact of classification grouping, as intra-G/P transitions contribute minimally ($0.21\%$, $0.06\%$, and $0.18\%$, respectively). Practically, these patterns indicate that PLMs explore a wider substitution landscape, whereas LLM-based architectures preferentially propose mutations that conserve fundamental physicochemical properties.

Both behaviors reflect plausible biological heuristics conditioned on available inputs. Sequence-based PLMs effectively identify substitutions compatible with evolutionary alignments and structural stability, which frequently transcend rigid residue classes. The conservative substitutions observed for standalone LLMs are consistent with general biochemical heuristics, although these results do not identify the source of their internal priors. AMixKit agents can additionally call PLMs through tools. However, neither prior explicitly captures assay-specific determinants (e.g., active-site geometry, interface kinetics, or conformational dynamics) that ultimately dictate measured phenotypes. Consequently, these trends reflect contrasting search strategies rather than an intrinsic hierarchy in model quality.

\subsubsection{Utilization of Multi-mutant Component Additivity}
\label{subsec:multi-mutant-differences}

\begin{figure}[t]
    \centering
    \includegraphics[width=\textwidth]{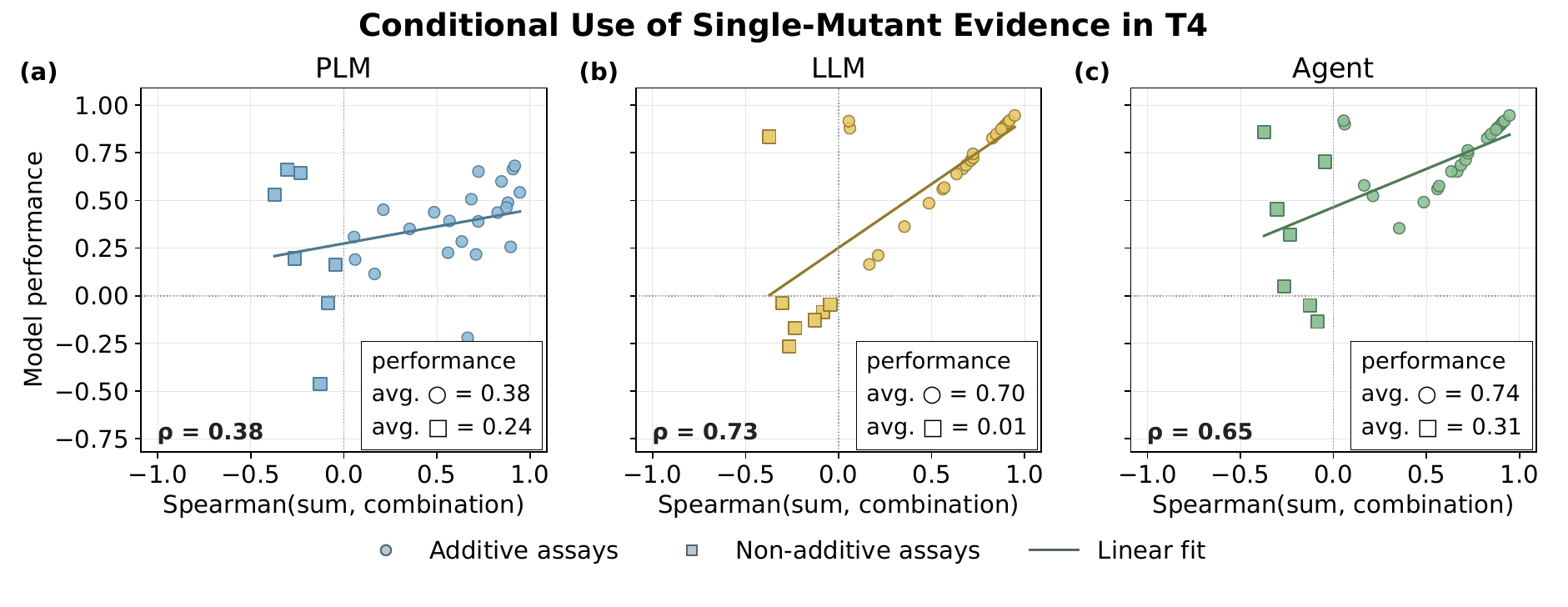}
  \caption{\textbf{Conditional utilization of single-mutant evidence in multi-mutant ranking.}
    Panels (a)--(c) show PLMs, LLMs and agents, respectively, with each point corresponding to one \ntfour assay.
    The horizontal axis is the Spearman correlation between the sum of measured component
    single-mutant fitness and the measured multi-mutant fitness. This is an information-transfer
    diagnostic, not a direct estimate of a mechanistic epistasis coefficient. \textbf{Circles} mark
    assays with $\rho_{\mathrm{sum,combination}}\geq0$ and \textbf{squares} mark assays below
    this operational threshold. The line marks a linear fit, while $\rho$ is the
    ranking correlation between the diagnostic and model performance.}
    \label{fig:pfarena-multimutant}
\end{figure}

For multi-mutant ranking tasks, the experimental context supplied to the model increases progressively. A key observation is that while PLMs, LLMs and agents exhibit comparable performance when single-mutant evidence is limited, LLMs and agents gain a distinct advantage as component-level measurements become available. Across \ntasktwo and \ntaskthree, the three model families achieve mean assay-level Spearman correlations of $0.271$ vs. $0.331$ (PLMs), $0.223$ vs. $0.311$ (LLMs), and $0.317$ vs. $0.383$ (agents). However, this performance gap widens dramatically in \ntaskfour—where measured single-mutant effects are explicitly provided—with LLMs and agents achieving $0.535$ and $0.635$, substantially outperforming PLMs at $0.349$.

This performance trajectory suggests that LLM-based architectures leverage individual component \dmsscore values through an additive heuristic. To test this hypothesis, we computed the sum of measured single-mutant \dmsscore values for each multi-mutant candidate in \ntfour and evaluated its correlation with actual combination fitness. As depicted in \Cref{fig:pfarena-multimutant}, a high correlation indicates that combination rankings can be reliably approximated via additive components, whereas a low correlation denotes poor transferability.

The predictive accuracy of LLMs ($\rho=0.733$, $P=6.12\times10^{-6}$) and agents ($\rho=0.649$, $P=1.39\times10^{-4}$) strongly correlates with the applicability of this additive heuristic, whereas PLMs display a much weaker coupling ($\rho=0.378$, $P=0.043$). On the non-additive regime of $\rho_{\mathrm{sum,combination}}<0$, comprising 7 of 29 assays, LLM performance drops sharply to $0.015$, compared to $0.242$ for PLMs and $0.314$ for agents. Conversely, on the high-transfer split, mean performance reaches $0.701$ for LLMs, $0.737$ for agents, but only $0.384$ for PLMs.

These findings reveal a conditional dependency on component evidence: LLMs and agents excel when combination fitness is well-approximated by additive single-mutant values, but LLMs collapse when this assumption fails. Although agentic workflows provide partial robustness in non-additive regimes, neither LLMs nor agents show evidence of modeling complex non-additive residue interactions. PLMs are less coupled to this diagnostic and therefore show smaller performance degradation, but not a general advantage on the low-transfer assays. Consequently, the dominance of LLM-based systems in \ntfour reflects effective utilization of transferable component evidence rather than a comprehensive understanding of residue coupling.

\subsubsection{Common Limitations across Single-mutant and Multi-mutant Tasks}
\label{subsec:shared-difficulties}

\begin{figure}[t]
    \centering
    \includegraphics[width=\textwidth]{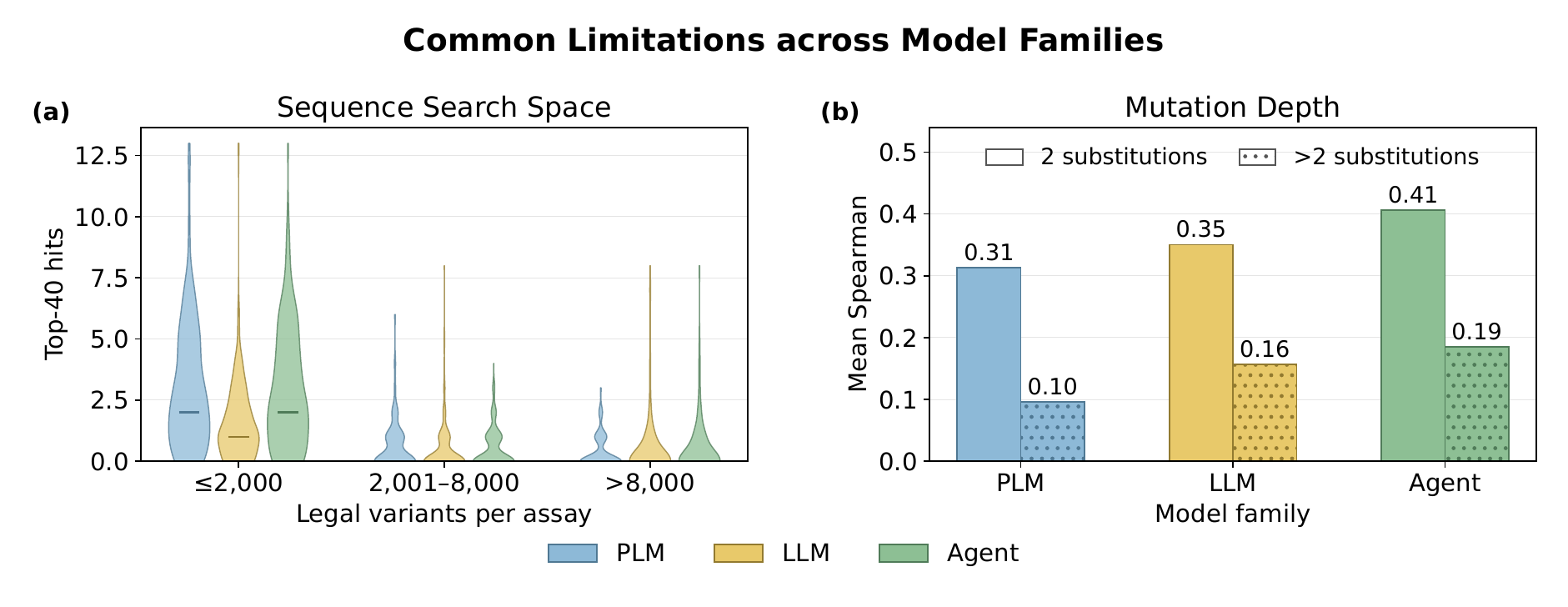}
    \caption{\textbf{Common limitations across model families.} \textbf{(a)} Sequence search space. Verified \ntone top-40 hit counts, stratified by the number of legal variants per assay into three bins ($\leq$2{,}000; 2{,}001--8{,}000; $>$8{,}000). Each violin shows the distribution of hit counts across assays for one family (PLMs, LLMs or agents), with a horizontal line marking the median; the three families are offset within each bin and share a common color coding. \textbf{(b)} Mutation depth. Mean assay-level Spearman ranking accuracy for double mutants (solid) versus candidates with more than two substitutions (hatched), shown per family with the mean value labeled above each bar. PLMs, LLMs and agents use the same family colors in both panels.}
    \label{fig:pfarena-shared}
\end{figure}

Whether through distinct strategies for selecting amino-acid mutations, the application of single-mutation fitness values, or the response mechanisms to additive effects, these differences ultimately reflect variations in the models' capabilities across diverse tasks.
Yet all models face the same two dilemmas: sensitivity to sequence length, and a decline in prediction accuracy for variants of higher-orders. 

\paragraph{High-dimensional Search Spaces}
Navigating the search space of a protein sequence presents a severe challenge.
In \ntaskone, a single assay features a median of $2{,}964$ valid single-point substitutions (ranging from $1{,}093$ to $22{,}536$). However, each model proposes only its top $40$ candidates, thereby sampling a mere $1.35\%$ of the median search space. As depicted in \hyperref[fig:pfarena-shared]{Figure~\ref*{fig:pfarena-shared}(a)}, the average number of top-40 hits in the single-mutant scenario of \ntone is only $1.21$, with even the highest-performing case recovering only $13$. Furthermore, this recovery rate degrades rapidly with increasing sequence length and search space size. This weak performance only marginally outperforms random sampling, and indicates that current models fall far short of truly mastering sequence space or providing reliable candidate fitness evaluations. Extending this task to multi-mutant proposal across full-length sequences will exacerbate this challenge, as the combinatorial search space expands exponentially.

\paragraph{High-order Combinations}
The mutation depth within a variant (i.e., the number of substituted sites per combination) significantly impacts model ranking performance.
Across the multi-mutant ranking tasks of \nttwo--\ntfour, ranking accuracy declines monotonically as mutation count increases ($\rho=-0.430$, $P=4.75\times10^{-9}$), as shown in \hyperref[fig:pfarena-shared]{Figure~\ref*{fig:pfarena-shared}(b)}. 
Specifically, in \ntaskfour, the mean Spearman correlation for LLMs drops from $0.669$ on $2$ mutations to $0.514$ for $3$--$4$ mutations, and down to $0.222$ for variants with over 5 mutated sites, demonstrating superior efficacy on lower-order combinations. PLMs exhibit a parallel reduction from $0.353$ to $0.236$ and $0.051$, while LLM-based agents also show decay from $0.674$ to $0.536$ and $0.386$. Notably, even when augmented with single-mutant fitness values—which substantially improve double-mutant ranking capabilities—models fail to generalize to higher-order variants involving complex epistatic interactions. This pervasive performance drop highlights the fundamental difficulty of modeling high-order combinations.

Together, these findings highlight how sequence dimension and mutation depth compound search pressure on computational models. 
Given how heavily performance degrades even within the controlled settings of \ntone--\ntfour, unconstrained multi-site generation—a common scenario in practical protein engineering—presents an exponentially higher barrier. Addressing real-world bio-design problems requires robust predictive capability across high-dimensional search spaces and higher-order combinations. This underscores the core motivation behind our task design: models must maintain reliable analytical and judgment capabilities when subjected to simultaneous scaling in sequence length and mutation depth. \textbf{\bench} reveals the shortcomings of existing models along these crucial dimensions.

\subsection{Dive into LLMs}
\label{sec:ai}

\subsubsection{Feedback-guided In-context Adaptation}
\label{sec:multiround_adaptation}

\begin{figure}[t]
    \centering
    \includegraphics[width=1.00\linewidth]{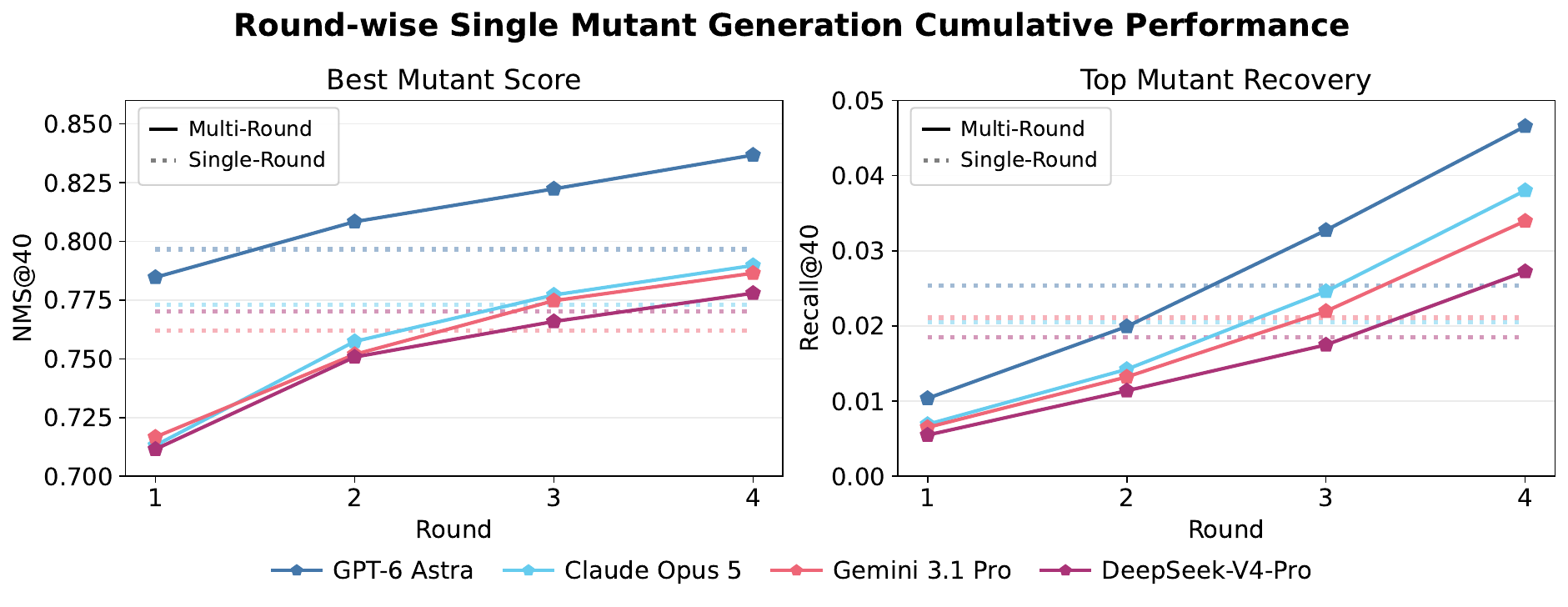}
    \caption{\textbf{Cumulative performance comparison of round-wise single mutant generation.} \textbf{Single-round} refers to the original \ntaskone setting, where all 40 mutations are proposed in a single pass, and is indicated with dotted horizontal lines. \textbf{Multi-round} refers to the results after a complete four-round feedback-guided generation protocol, with 10 mutations proposed each round, where proposed mutants with an assay-recorded \texttt{DMS\_score} are used as scored context for the next model pass, and is depicted in solid lines.}
    \label{fig:roundwise_t1_results}
\end{figure}

To emulate a sequential wet-laboratory campaign, in which mutation fitness is measured in batches and later candidates are selected using earlier outcomes, we extended the single-round \ntaskone into a four-round adaptive search. This experiment tests whether assay-specific fitness measurements supplied in context help models refine subsequent generations and identify higher-fitness mutants within a fixed screening budget. In each round, the model proposed 10 single mutants unselected in earlier rounds. For the next round, the ground-truth \dmsscore fitness of all previously proposed mutants was added to the prompt as in-context feedback, if the fitness value of the mutant is available. Both settings therefore used the same total budget of 40 proposed mutations, but only the multi-round setting allowed later generations to condition on intermediate fitness measurements.

As shown in \Cref{fig:roundwise_t1_results}, the multi-round protocol shows monotonic improvements in both NMS@$40$ and Recall@$40$ across rounds for every model. Relative to single-round generation, the final Round-4 endpoint NMS@$40$ increased from $0.7966$ to $0.8367$ for GPT-6 Astra, highest across all models, and increased by $0.0168$ for Claude Opus 5, $0.0244$ for Gemini 3.1 Pro, and $0.0076$ for DeepSeek-V4-Pro. Recall@$40$ increased by $0.0211$, $0.0175$, $0.0128$, and $0.0087$ respectively, with GPT-6 Astra achieving the highest recovery of $0.0465$. Averaged across the four models, multi-round feedback-guided generation improved NMS@$40$ by $0.0222$ and Recall@$40$ by $0.0150$. All Round-4 results exceeded the random baseline on both metrics, whereas three of the four single-round models remained below random sampling in NMS@$40$.

Across rounds, GPT-6 Astra exceeded its single-round NMS@$40$ baseline after two rounds, Claude Opus 5 and Gemini 3.1 Pro after three rounds, and DeepSeek-V4-Pro after four rounds. The first three models exceeded their single-round baselines in Recall@$40$ after three rounds, whereas DeepSeek-V4-Pro after the fourth. As the cumulative candidate budget increases across rounds, the earlier rounds alone do not establish the value of feedback; the decisive comparison is the Round-4 endpoint against single-round generation. The consistent performance gains demonstrate the effectiveness of feedback-guided in-context adaptation, particularly for GPT-6 Astra, which substantially surpassed its already strong single-round baseline, highlighting its capability in knowledge-based work tasks.
Notably, LLMs frequently revisited previously high-scoring sites with alternative amino-acid mutations, indicating that further exploration of promising positions in subsequent rounds may help identify mutants of higher fitness.

\subsubsection{Uncertainty Estimation}
\label{sec:ue}

\begin{figure}[t]
    \centering
    \includegraphics[width=1.00\linewidth]{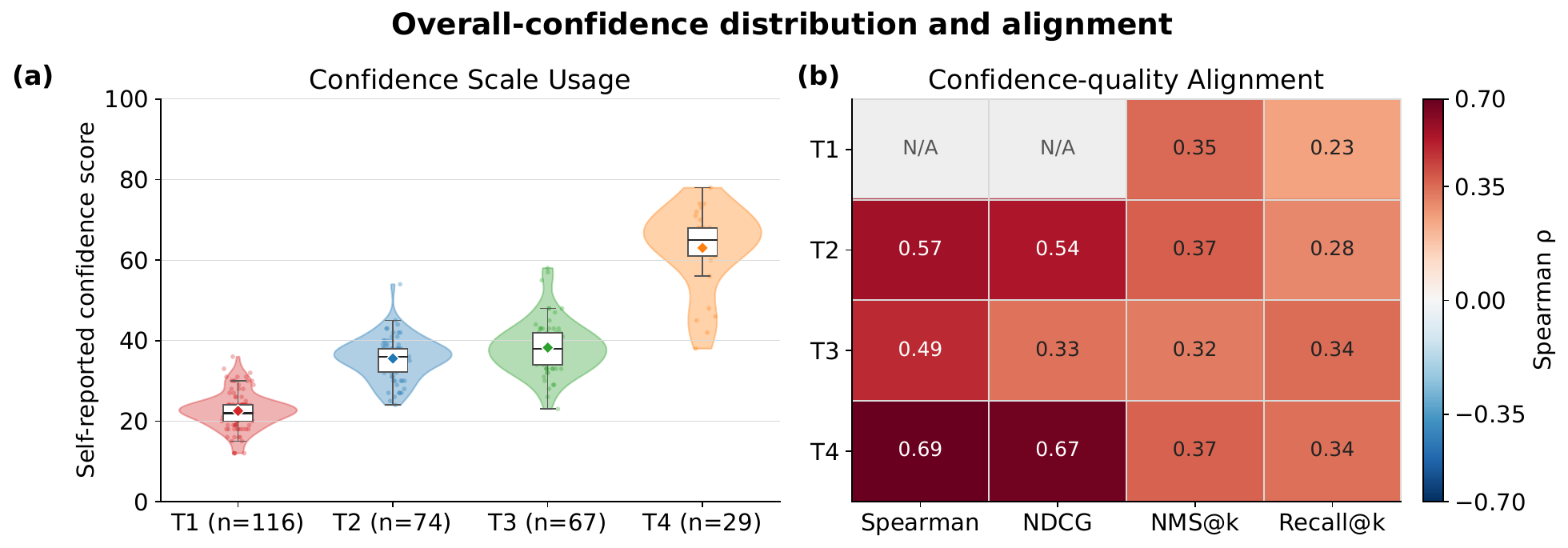}
    \caption{\textbf{Confidence distribution and quality alignment.} \textbf{(a)} GPT-6 Astra’s self-reported confidence across \ntone--\ntfour. \textbf{(b)} Within-setting Spearman correlations between confidence and ranking-quality metrics.}
    \label{fig:confidence1}
\end{figure}

We evaluated self-reported confidence from GPT-6 Astra to examine whether LLMs can estimate their own predictive uncertainty. For queries in \ntone--\ntfour, the model was prompted to assign an overall confidence score from $0$ to $100$ reflecting its confidence in identifying and prioritizing high-fitness mutations. Following five API policy refusals and two safety exclusions in \ntone, confidence was available for $116$, $74$, $67$, and $29$ queries in \ntone--\ntfour, respectively. We treated this score as an inverse measure of relative query-level uncertainty and assessed its association with predictive performance using Spearman correlation. The confidence elicitation prompt used in our experiments is provided in Appendix~\ref{app:confidence-prompt}.

As shown in \hyperref[fig:confidence1]{Figure~\ref*{fig:confidence1}(a)}, confidence varied substantially across settings, with mean scores of $22.5$, $35.6$, $38.3$, and $63.0$ for \ntone--\ntfour, respectively, and no response reaching $90$. \ntaskone elicited relatively conservative judgments, whereas \ntaskfour produced a relatively high-confidence distribution. These absolute values should not be interpreted as success probabilities or directly compared across settings, which differ in data and evaluation objectives. The more meaningful test is whether confidence distinguishes higher- from lower-quality predictions within each setting.

Self-reported confidence provides a partial, task- and context-dependent estimate of ranking reliability. \hyperref[fig:confidence1]{Figure~\ref*{fig:confidence1}(b)} shows positive evidence in \ntaskthree, where confidence aligned with all four quality dimensions ($\rho\approx0.32$--$0.49$) after multiple-testing correction. In \ntasktwo, confidence also aligned with all four dimensions ($\rho\approx0.28$--$0.57$), with stronger associations for global ordering than top-five selection. \ntone showed weak-to-moderate associations with its top-40 metrics ($\rho=0.35$ for NMS@$40$ and $0.23$ for Recall@$40$), which remained significant after filtering dirty outputs. \ntfour showed the largest correlations for full-list Spearman and NDCG quality ($\rho=0.69$ and $0.67$), whereas neither top-five metric passed multiple-testing correction. Assay-level examples reinforce this limitation: one protein-stability query in \ntfour received high confidence ($78$) despite limited top-five recovery (Recall@$5$=$0.2$), whereas a fluorescence query received lower confidence ($45$) but substantially better recovery (Recall@$5$=$0.8$). These cases show that confidence does not uniformly track top-candidate recovery, even when it aligns with global ranking quality. It can therefore support coarse query triage within a fixed task, but should not be treated as a generally reliable or cross-assay uncertainty estimate.

\begin{figure}[t]
    \centering
    \includegraphics[width=1.00\linewidth]{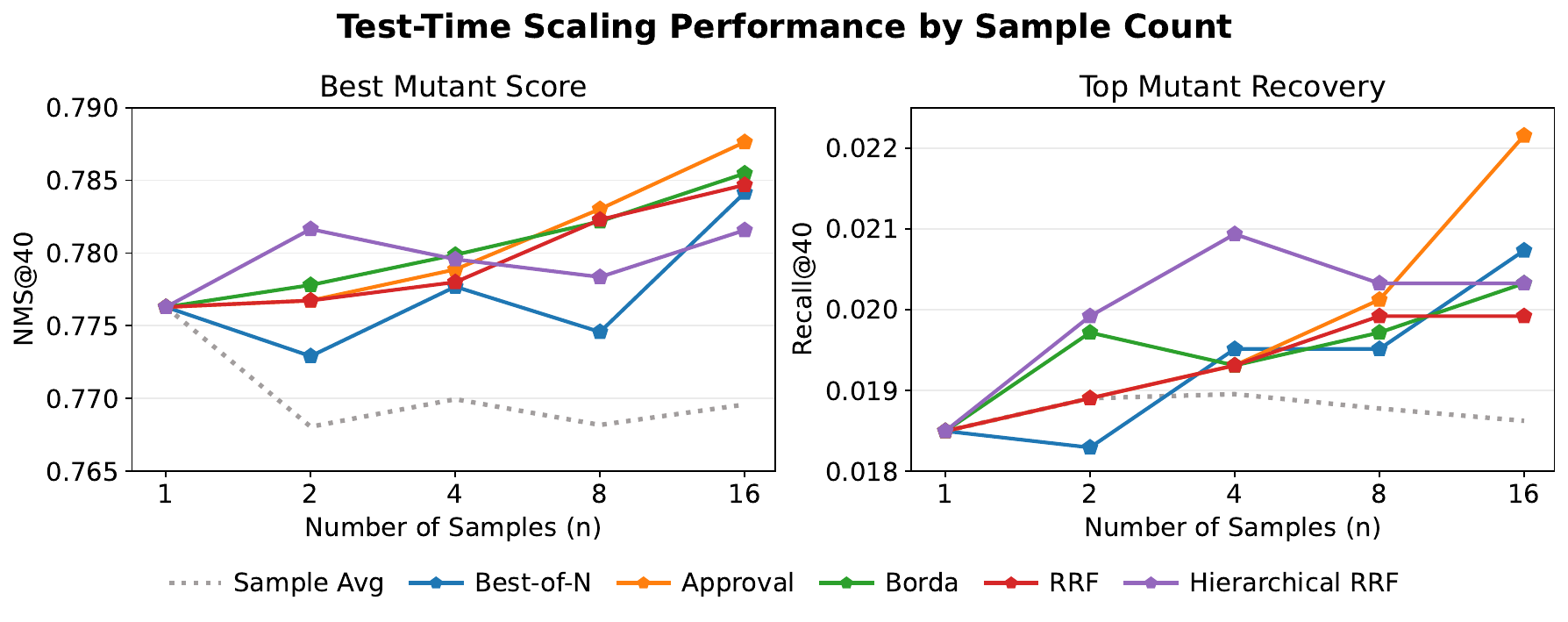}
    \caption{\textbf{Test-time scaling performance across sample counts.} The panels show NMS@$40$ (left) and Recall@$40$ (right) as the number of sampled mutation lists increases, with the dotted line representing the sample average.}
    \label{fig:tts_scaling}
\end{figure}

\begin{table}[t]
\centering
\small
\renewcommand{\arraystretch}{1.05}
\setlength{\tabcolsep}{4.5pt}

\begin{tabular}{
    l
    l
    S[table-format=1.4]
    S[table-format=1.4]
    S[table-format=1.4]
}

\toprule
\textbf{Setting}
    & \textbf{Method}
    & \multicolumn{1}{c}{\textbf{NMS@$40$ $\uparrow$}}
    & \multicolumn{1}{c}{\textbf{Recall@$40$ $\uparrow$}}
    & \multicolumn{1}{c}{\textbf{Invalid per Assay} $\downarrow$} \\
\midrule
\addlinespace[0.5em]

\multirow{2}{*}{Vanilla}
    & Sample Avg
    & 0.7696 & 0.0186 & 2.2805 \\
    & RRF
    & {\bfseries 0.7847} & {\bfseries 0.0199} & {\bfseries 1.4472} \\

\addlinespace[0.2em]
\midrule
\addlinespace[0.5em]

\multirow{2}{*}{Clean}
    & Sample Avg
    & 0.7771 & 0.0189 & {\bfseries 1.0531} \\
    & RRF
    & {\bfseries 0.7845} & {\bfseries 0.0197} & 1.1057 \\

\addlinespace[0.2em]
\midrule
\addlinespace[0.5em]

\multirow{2}{*}{In-Assay}
    & Sample Avg
    & 0.7745 & 0.0172 & 0.0000 \\
    & RRF
    & {\bfseries 0.7842} & {\bfseries 0.0191} & 0.0000 \\
\addlinespace[0.2em]
\bottomrule
\end{tabular}

\caption{Validity-controlled RRF analysis at $n=16$. Clean requires 40 unique and sequence-valid mutations. In-Assay further requires all mutations to appear in the assay's DMS table.}
\label{tab:TTS_ablation}
\end{table}

\subsubsection{Test-time Scaling}
\label{sec:tts}

To test whether additional inference-time computation improves single-mutation generation without experimental feedback, we independently sampled 16 responses from DeepSeek-V4-Pro using exactly the same input for each \ntaskone query. We extracted a ranked top-40 mutation list from each response and applied aggregation to nested prefixes with $n\in\{1,2,4,8,16\}$. Aggregation used no ground-truth \dmsscore, candidate labels, or evaluator metrics, and the mean performance of the corresponding raw samples at each prefix size served as the Sample Avg baseline.

We compared five TTS aggregation strategies. \textbf{Best-of-N} selected the complete sampled ranking most consistent with the other samples. \textbf{Approval Voting} prioritized mutations that appeared repeatedly across samples, whereas \textbf{Borda Fusion} and \textbf{Reciprocal Rank Fusion (RRF)} additionally incorporated within-sample rank using linear and reciprocal weighting, respectively. \textbf{Hierarchical RRF} further pooled evidence across different mutations at the same residue position. In the main scaling experiments, all methods operated directly on the extracted candidate strings without explicit sequence-validity or DMS-membership filtering. Complete method details and hyperparameters used in our experiments are provided in \Cref{app:tts}.

As shown in \Cref{fig:tts_scaling}, increasing sample count generally improved performance. Approval scaled most consistently, reaching $0.7876$ NMS@$40$ and $0.0222$ Recall@$40$ at $n=16$, versus $0.7696$ and $0.0186$ for Sample Avg. Borda and RRF also improved with more samples, while Best-of-N was less stable and Hierarchical RRF mainly benefited small $n$. All five methods outperformed Sample Avg at $n=16$.

One source of improvement was invalid-candidate suppression. LLM-proposed mutations may be \textit{locally invalid} due to sequence inconsistency or formatting errors, or \textit{assay invalid} when they do not appear in the assay's DMS table. In the main setting, TTS aggregation substantially suppressed invalid outputs. As shown in \Cref{tab:TTS_ablation}, RRF reduced invalid predictions from $2.2805$ to $1.4472$, without explicit validity filtering. To determine whether this fully explains the gain, we further evaluated \textbf{Clean} and \textbf{In-Assay} settings. Clean retained only samples with 40 unique and locally valid candidates, while In-Assay further required all candidates to appear in the assay's DMS table. Under \textbf{Clean}, RRF improved NMS from $0.7771$ to $0.7845$ and recall from $0.0189$ to $0.0197$, despite slightly more assay-absent predictions ($1.0531$ to $1.1057$). Under \textbf{In-Assay}, where both methods had zero invalid predictions, RRF still improved NMS from $0.7745$ to $0.7842$ and recall from $0.0172$ to $0.0191$. Thus, although invalid-candidate suppression significantly contributes to the TTS gain, it does not fully account for the improvement; even when validity is controlled, aggregation helps identify and prioritize higher-quality valid candidates across repeated samples from the same query.

Despite the gains from TTS at larger sample counts, a substantial oracle gap remains. At $n=16$, best-sample oracles reached $0.8474$ NMS@$40$ and $0.0461$ Recall@$40$, compared with $0.7876$ and $0.0222$ for the best practical TTS method. This gap suggests that repeated sampling can already produce substantially better predictions, but reliably identifying them without experimental feedback remains challenging.

%% file: sections/02related.tex
\section{Related Work}
\label{sec:related_benchmarks}

\paragraph{Protein Fitness Benchmarks}
Protein fitness benchmarks have established standardized evaluation of computational models for predicting experimentally measured mutation effects. FLIP~\cite{dallago2021flip} evaluates fitness-landscape inference under low-resource and extrapolative generalization settings, while FLIP2~\cite{didi2026flip2} extends this framework to unseen mutation identities, positions, mutation counts, wild-type proteins, and high-fitness regions. ProteinGym~\cite{notin2023proteingym} further enables large-scale evaluation of zero-shot and supervised mutation-effect predictors across diverse DMS assays and clinical labels. Together, these benchmarks characterize how accurately PLMs score and rank predefined mutants across heterogeneous fitness landscapes.

Recent benchmarks have extended protein fitness evaluation to general-purpose LLMs and LLM-based agents. ProteinGym-LLM~\cite{arora2026proteingymllm} evaluates whether LLMs can rank a fixed set of protein mutants from the wild-type sequence and assay description, providing a direct assessment of LLM-based mutation prioritization. BioDesignBench~\cite{kim2026biodesignbench} evaluates tool-using LLM agents across expert-curated protein-design workflows, with an emphasis on candidate generation, tool use, and iterative evaluation. These studies extend protein engineering benchmarks from specialized fitness predictors to general-purpose reasoning and agentic design systems.

Despite these advances, existing benchmarks typically focus on either fixed-list mutant ranking or end-to-end protein-design workflows, leaving systematic comparison across different mutation-discovery settings underexplored. \textbf{\bench} addresses this gap by evaluating PLMs, general-purpose LLMs, and LLM-based agents through one single-mutant generation task (\ntone) and three multi-mutant ranking tasks under measurement-free (\nttwo), anchor-informed (\ntthree), and single-mutant-informed (\ntfour) conditions, applying consistent task interfaces and evaluation criteria across all model families.

\paragraph{Protein Fitness Methods}
Target-specific supervised methods provide the most direct approach to protein fitness prediction by learning sequence--fitness relationships from experimentally measured mutants. The Low-$N$ framework~\cite{biswas2021lown} uses pretrained UniRep representations and as few as 24 assayed mutants to guide protein engineering, while Hsu et al.~\cite{hsu2022fitness} combine site-specific amino-acid features with evolutionary density estimates to predict fitness from limited measurements. Because these methods are trained against the phenotype of interest, they can directly adapt to the assay-specific objective, but their performance depends on the number, diversity, and sequence-space coverage of available labels.

Zero-shot PLMs avoid target-specific training by deriving mutation scores from general biological priors learned from sequence, evolutionary, and structural data. Evolutionary models such as EVmutation~\cite{hopf2017evmutation}, DeepSequence~\cite{riesselman2018deepsequence}, and EVE~\cite{frazer2021eve} infer family-specific constraints from homologous sequences. Protein language models, including ESM-1v~\cite{meier2021zeroshot}, ESM-2~\cite{lin2023evolutionary}, ProGen2~\cite{nijkamp2023progen2}, Tranception~\cite{notin2022tranception}, and DPLM-Evo~\cite{wang2026towards} instead learn sequence compatibility across large protein corpora. Structure-aware models such as ProSST~\cite{li2024prosst}, S3F~\cite{zhang2024s3f}, and VenusREM~\cite{tan2025venusrem} further incorporate geometric and evolutionary information. These models can score mutations without assay-specific labels, but their scores primarily reflect general protein plausibility rather than the phenotype and experimental evidence associated with a particular engineering objective.

General-purpose LLMs and scientific agents have also shown substantial potential for protein mutation discovery. Successive Claude releases have demonstrated continued progress in protein mutation-effect prediction, with Claude Opus 5 further improving upon previous generations on ProteinGym Hard~\cite{anthropic2026claude5}. LLMs have also been incorporated into budget-constrained protein optimization procedures~\cite{wang2025llmoptimizer}, suggesting that their biological knowledge can support mutation selection beyond direct fitness scoring. Scientific agents further extend these capabilities through retrieval and tool use: ProtAgents~\cite{ghafarollahi2024protagents} coordinates specialized agents for protein analysis and design, while Biomni~\cite{huang2025biomni} integrates biomedical databases, scientific software, and code execution. These developments motivate systematic evaluation of whether LLMs and agents can convert assay context, experimental evidence, and specialized protein-model outputs into effective mutation priorities.

Given the rapidly expanding set of PLMs, LLMs, and scientific agents, our evaluation focuses on representative methods spanning sequence-based, structure-based, evolution-based, and tool-based paradigms.

%% file: sections/06conclusion.tex
\section{Conclusion}
\label{sec:conclusion}

\textbf{\bench} provides a unified evaluation of PLMs, LLMs, and LLM-based agents across realistic protein modification decisions. Taken together, our results show that the preferred model family depends on the task interface. Protein-specific representations are particularly effective for open-ended single-mutant search, whereas general-purpose reasoning and tool use become more useful for ranking mutation combinations, especially when measured single-mutant context is available. This complementarity does not eliminate shared bottlenecks: high-fitness candidate recovery remains sparse under limited prediction budgets, and ranking becomes substantially less reliable beyond double mutants. Our additional analyses show that feedback-guided adaptation and test-time scaling can improve LLM-based inference. We release our code and benchmark suite to support reproducible progress on these open problems.

\paragraph{Limitations}
Despite its broad coverage, \textbf{\bench} has several limitations.

\begin{itemize}
    \item \textbf{Potential label leakage.}
    All assays in \textbf{\bench} are derived from publicly available datasets and publications. Their sequences, mutation labels, or experimental results may therefore have appeared in the pretraining corpora of the evaluated LLMs, making it impossible to fully exclude memorization or other forms of data contamination that could influence the evaluation results.

    \item \textbf{Stochasticity and single-run evaluation.}
    Because evaluating frontier LLMs and tool-augmented agents is computationally and financially expensive, each LLM or agent configuration was run only once for main evaluation. The reported results may therefore depend on sampling randomness.

    \item \textbf{Missing outputs, refusals, and comparison fairness.}
    Some LLMs, particularly proprietary systems such as GPT and Claude, refuse to answer a subset of assays because of their safety policies. To retain a fixed evaluation set and avoid selectively excluding difficult cases, we replace such outputs with a deterministic random baseline result, as detailed in \Cref{app:llm_missing_assays}. This protocol measures the end-to-end usability of each deployed system, but it also conflates underlying protein-reasoning ability with provider-specific refusal policies and output reliability.

    \item \textbf{Retrospective rather than prospective evaluation.}
    \textbf{\bench} is constructed from previously measured assays. Improvements in benchmark metrics consequently do not establish higher prospective wet-lab hit rates or guarantee that the selected mutations will succeed in a new experimental campaign. Prospective validation will be required to determine whether the observed performance gains translate into practical reductions in experimental cost and design cycles.

    \item \textbf{Dataset coverage and heterogeneity.}
    The assays vary in experimental protocol, measurement noise, sequence coverage, candidate-library construction, and phenotype definition. Harmonization and quality control reduce but cannot eliminate these differences. Moreover, publicly available datasets may overrepresent well-studied proteins, and assay types that are convenient to measure, while underrepresenting negative results, rare protein families, and complex cellular or organism-level phenotypes. The conclusions may therefore not generalize to all protein-engineering settings.
\end{itemize}

%% file: sections/contributions.tex
\newpage
\section*{Contributions}

\textbf{Project Lead}

Yawen Ouyang$^{1,2}$

\textbf{Co-first Authors}

Yawen Ouyang$^{1,2}$, Xinbo Zhang$^{1,2}$, Ziyuan Ma$^{1,4}$

\textbf{Task Design and Data Collection}

Ziyuan Ma$^{1,4}$, Xinbo Zhang$^{1,2}$, Yawen Ouyang$^{1,2}$

\textbf{Main Results}

Yawen Ouyang$^{1,2}$, Yixin Wu$^{1,2}$, Wenbin Liao$^{1,5}$, Xinbo Zhang$^{1,2}$

\textbf{Analysis}

Ziyuan Ma$^{1,4}$, Yixin Wu$^{1,2}$, Wenjie Li$^{1}$, Feiran Zhang$^{1,6}$, Yawen Ouyang$^{1,2}$

\textbf{Other Contributors}

Lihao Wang$^{1,2}$, Hao Wang$^{2,3}$, Xiaoqing Zheng$^{6}$, Xuefeng Yan$^{5}$, Lei Bai$^{1}$, Ya-Qin Zhang$^{3}$, Shuyi Zhang$^{4}$, Wei-Ying Ma$^{3,7}$, Dahua Lin$^{1}$, Bowen Zhou$^{1}$

\textbf{Correspondence}

Hao Zhou$^{1,2,3}$

\subsection*{Affiliation}

$^1$Shanghai Artificial Intelligence Laboratory

$^2$Generative Symbolic Intelligence Lab (GenSI), Tsinghua University

$^3$Institute for AI Industry Research (AIR), Tsinghua University

$^4$School of Pharmaceutical Sciences, Tsinghua University

$^5$School of Information Science and Engineering, East China University of Science and Technology

$^6$College of Computer Science and Artificial Intelligence, Fudan University

$^7$City University of Hong Kong

\section*{Acknowledgments}

This work is supported by Shanghai Artificial Intelligence Laboratory and NSFC (Grant No. 62406170). 

%% file: sections/appendix.tex
\section{Evaluation Templates}
\label{app:prompt_templates}

To support standardized and reproducible evaluation, we provide the prompt templates used for the evaluation of general-purpose language models on \textbf{\bench}. Agentic models use similar core templates, supplemented with additional instructions for tool calling.

\subsection{Single-mutant Generation}

Task \ntaskone provides the language model with the \texttt{wildtype\_sequence} of the evaluated protein, its \texttt{sequence\_length}, and its UniProt identifier (\texttt{uniprot\_id}, e.g., ``P11413''). A concise description of its experimental assay context is also passed to the model, including the \texttt{primary\_task\_class} (e.g., ``activity\_function''), \texttt{fitness\_type} (e.g., ``enzymatic\_activity''), and \texttt{assay\_readout\_subclass} (e.g., ``activity\_proxy''). 
The model is instructed to generate a list of plausible mutations and rank them according to their expected fitness based on its own internal knowledge.

\begin{promptbox}{Single-mutant Generation Prompt Template}
You are an expert protein engineer and computational biologist specializing in deep mutational scanning (DMS) and mutation-effect prediction.

### TASK GOAL
Given a wild-type protein sequence and its experimental assay context, predict the **top 40 single point mutations** (WTposMUT format) that optimize the target fitness metric, ordered from highest expected fitness to lowest expected fitness.

### REASONING & EVIDENCE BOUNDARIES
1. **Biochemical Deductions**: Analyze residue chemistry, conservation, secondary structure propensities, steric packing, hydrophobic cores, electrostatic interactions, and sequence motifs within the supplied wild-type sequence.
2. **Assay Alignment**: Align every ranked mutation strictly with the supplied assay readout. For example, if evaluating stability/abundance, prioritize mutations that improve hydrophobic packing or thermostability without disrupting necessary structural dynamics.
3. **No Hallucination**: Do NOT invent or claim specific numerical model scores, experimental PDB coordinates, literature measurements, or alignments that are not logically derivable from sequence biochemistry.
4. **Single-Response Constraint**: You have no external tools, browsing, or follow-up turns. Complete the analysis in this single response.

### STRICT MUTATION & FORMAT CONSTRAINTS
1. **Ranking Size**: The output list MUST contain **exactly 40 mutations**, ordered from best to worst.
2. **Format**: Every mutant MUST be represented in standard 1-indexed `WTposMUT` format (e.g., `H24R`, `A15V`).
3. **Alphabet**: Both `WT` and `MUT` must be standard 20 amino acid single-letter codes: `A, C, D, E, F, G, H, I, K, L, M, N, P, Q, R, S, T, V, W, Y`.
4. **Uniqueness**: Every mutation string MUST appear exactly once (no duplicates, no omissions within your top-40 list).
5. **Strict Validation**:
   - Every mutation position follows `1 <= position <= Length`.
   - `WT` MUST strictly match the character at `wildtype_sequence[position - 1]`.
   - `MUT` MUST be strictly different from `WT` (no synonymous/no-op mutations).
6. **Search Space**: Consider all valid single amino acid substitutions across the full wild-type sequence, then return only the top 40.
7. **Prohibited Formats**: Multi-site mutations, insertions (`ins`), deletions (`del`), stop codons (`*`), HGVS notations, or numerical confidence scores.

---

### OUTPUT FORMAT
Return a **valid JSON object ONLY** with NO markdown code block wrappers, prefix, or conversational text. Use the following exact JSON schema:

{
  "ranking": [
    "<best mutant>",
    "<second-best mutant>",
    "..."
  ]
}

---

### INSTANCE DATA
1. ASSAY CONTEXT & TARGET METRICS
- **UniProt ID**: {uniprot_id}
- **Primary Task Class**: {primary_task_class}
- **Fitness Metric Type**: {fitness_type}
- **Assay Readout Subclass**: {readout_subclass}

2. INPUT WILD-TYPE SEQUENCE (Length: {sequence_length})
`{wildtype_sequence}`
\end{promptbox}

\subsection{Multi-mutant Ranking}

For the mutation-ranking tasks of \nttwo--\ntfour, the protein and assay metadata included in \ntone is similarly provided. Additionally, the model receives the total amount of candidates to rank \texttt{num\_candidates}, and a shuffled list of \texttt{candidate\_mutants}, each specified with the original amino-acid and position in the wild-type sequence followed by the mutant, in a ``WTposMUT'' format like ``\texttt{E19H}''.

\paragraph{Measurement-free Ranking}

No additional experimental measurements are provided for \ntasktwo. The model ranks the candidate mutants using only the shared information on the protein and assay context.

\begin{promptbox}{Measurement-free Multi-mutant Ranking Prompt Template}
You are an expert protein engineer and computational biologist specializing in deep mutational scanning (DMS) and mutation-effect prediction.

### TASK GOAL
Given a wild-type protein sequence, its experimental assay context, and a specific list of candidate multi-mutations, **rank all candidate multi-mutations from best to worst** according to their expected target fitness metric.

### REASONING & EVIDENCE BOUNDARIES
1. **Biochemical Deductions**: Compare the candidates based on residue chemistry, conservation, secondary structure propensities, steric packing, hydrophobic cores, electrostatic interactions, and sequence motifs within the supplied wild-type sequence.
2. **Assay Alignment**: Evaluate relative effects of these specific substitutions on the supplied assay readout. Rank mutations that better preserve or enhance structural/functional requirements above those that introduce severe clashes, charge mismatch, or instability.
3. **No Hallucination**: Do NOT invent or claim specific numerical model scores, experimental PDB coordinates, literature measurements, or alignments that are not logically derivable from sequence biochemistry.
4. **Single-Response Constraint**: You have no external tools, browsing, or follow-up turns. Complete the ranking in this single response.

### STRICT MUTATION & FORMAT CONSTRAINTS
1. **Closed Set Principle**: You MUST ONLY rank the mutations provided in the list above. Do NOT introduce new mutations, insertions, deletions, or wild-type strings.
2. **Multi-Mutation Candidates**: A candidate contains multiple mutations joined by + (e.g., K23A+A40P+T52S). When ranking multi-mutation candidates, account for their combined effects.
3. **Exact Copy & Completeness**:
   - The output list MUST contain **exactly `total_candidates` items**.
   - Every candidate from the input list MUST appear **exactly once** (no duplicates, no omissions).
   - Each mutation string MUST be copied **exactly as provided**.
4. **Strict Validation**:
   - Every mutation position follows `1 <= position <= Length`.
   - `WT` MUST strictly match the character at `wildtype_sequence[position - 1]`.
   - `MUT` MUST be strictly different from `WT` (no synonymous/no-op mutations).

---

### OUTPUT FORMAT
Return a **valid JSON object ONLY** with NO markdown code block wrappers, prefix, or conversational text. Use the following exact JSON schema:

{
  "ranking": [
    "<best mutant copied exactly from the provided list>",
    "<second-best mutant copied exactly from the provided list>",
    "..."
  ]
}

---

### INSTANCE DATA
1. ASSAY CONTEXT & TARGET METRICS
- **UniProt ID**: {uniprot_id}
- **Primary Task Class**: {primary_task_class}
- **Fitness Metric Type**: {fitness_type}
- **Assay Readout Subclass**: {readout_subclass}

2. INPUT WILD-TYPE SEQUENCE (Length: {sequence_length})
`{wildtype_sequence}`

3. CANDIDATE MUTATIONS TO RANK (Total Candidates: {num_candidates})
{candidate_mutants}
\end{promptbox}

\paragraph{Anchor-informed Ranking}

In task \ntaskthree, the model additionally receives an \texttt{anchor\_mutant} shared by all candidates and its ground-truth \texttt{anchor\_DMS\_score}. This measurement provides a common experimental reference for ranking the candidate mutations.

\begin{promptbox}{Anchor-informed Multi-mutant Ranking Prompt Template}
You are an expert protein engineer and computational biologist specializing in deep mutational scanning (DMS) and mutation-effect prediction.

### TASK GOAL
Given a wild-type protein sequence, its experimental assay context, and a specific list of candidate multi-mutations, **rank all candidate multi-mutations from best to worst** according to their expected target fitness metric.
<<The ground-truth DMS score of an anchor mutant appearing in every candidate, which may be single-site or multi-site, will also be provided.>>

### REASONING & EVIDENCE BOUNDARIES
1. **Biochemical Deductions**: Compare the candidates based on residue chemistry, conservation, secondary structure propensities, steric packing, hydrophobic cores, electrostatic interactions, and sequence motifs within the supplied wild-type sequence.
2. **Assay Alignment**: Evaluate relative effects of these specific substitutions on the supplied assay readout. Rank mutations that better preserve or enhance structural/functional requirements above those that introduce severe clashes, charge mismatch, or instability.
3. **No Hallucination**: Do NOT invent or claim specific numerical model scores, experimental PDB coordinates, literature measurements, or alignments that are not logically derivable from sequence biochemistry.
4. **Single-Response Constraint**: You have no external tools, browsing, or follow-up turns. Complete the ranking in this single response.

### STRICT MUTATION & FORMAT CONSTRAINTS
1. **Closed Set Principle**: You MUST ONLY rank the mutations provided in the list above. Do NOT introduce new mutations, insertions, deletions, or wild-type strings.
2. **Multi-Mutation Candidates**: A candidate contains multiple mutations joined by + (e.g., K23A+A40P+T52S)<<, with the anchor mutant included>>. When ranking multi-mutation candidates, account for their combined effects.
3. **Exact Copy & Completeness**:
   - The output list MUST contain **exactly `total_candidates` items**.
   - Every candidate from the input list MUST appear **exactly once** (no duplicates, no omissions).
   - Each mutation string MUST be copied **exactly as provided**.
4. **Strict Validation**:
   - Every mutation position follows `1 <= position <= Length`.
   - `WT` MUST strictly match the character at `wildtype_sequence[position - 1]`.
   - `MUT` MUST be strictly different from `WT` (no synonymous/no-op mutations).

---

### OUTPUT FORMAT
Return a **valid JSON object ONLY** with NO markdown code block wrappers, prefix, or conversational text. Use the following exact JSON schema:

{
  "ranking": [
    "<best mutant copied exactly from the provided list>",
    "<second-best mutant copied exactly from the provided list>",
    "..."
  ]
}

---

### INSTANCE DATA
1. ASSAY CONTEXT & TARGET METRICS
- **UniProt ID**: {uniprot_id}
- **Primary Task Class**: {primary_task_class}
- **Fitness Metric Type**: {fitness_type}
- **Assay Readout Subclass**: {readout_subclass}

2. INPUT WILD-TYPE SEQUENCE (Length: {sequence_length})
`{wildtype_sequence}`

<<3. ANCHOR MUTANT CONTEXT
- **Anchor Mutant**: {anchor_mutant}
- **Anchor DMS Score**: {anchor_DMS_score} >>

4. CANDIDATE MUTATIONS TO RANK (Total Candidates: {num_candidates})
{candidate_mutants}
\end{promptbox}

\paragraph{Single-mutant-informed Ranking}

Task \ntaskfour additionally provides the ground-truth fitness scores of every component mutation appearing in the candidates as \texttt{single\_mutant\_DMS\_score}. 
Each mutation--score pair is formatted as ``\texttt{<mutant>:\,<dms\_score>}'', and the pairs are joined by newline characters (``\texttt{\textbackslash n}'').
These measurements provide mutation-level evidence for ranking the multi-mutant combinations.

\begin{promptbox}{Single-mutant-informed Multi-mutant Ranking Prompt Template}
You are an expert protein engineer and computational biologist specializing in deep mutational scanning (DMS) and mutation-effect prediction.

### TASK GOAL
Given a wild-type protein sequence, its experimental assay context, and a specific list of candidate multi-mutations, **rank all candidate multi-mutations from best to worst** according to their expected target fitness metric.
<<The ground-truth single-mutant DMS scores of every component appearing in any candidate will also be provided.>>

### REASONING & EVIDENCE BOUNDARIES
1. **Biochemical Deductions**: Compare the candidates based on residue chemistry, conservation, secondary structure propensities, steric packing, hydrophobic cores, electrostatic interactions, and sequence motifs within the supplied wild-type sequence.
2. **Assay Alignment**: Evaluate relative effects of these specific substitutions on the supplied assay readout. Rank mutations that better preserve or enhance structural/functional requirements above those that introduce severe clashes, charge mismatch, or instability.
3. **No Hallucination**: Do NOT invent or claim specific numerical model scores, experimental PDB coordinates, literature measurements, or alignments that are not logically derivable from sequence biochemistry.
4. **Single-Response Constraint**: You have no external tools, browsing, or follow-up turns. Complete the ranking in this single response.

### STRICT MUTATION & FORMAT CONSTRAINTS
1. **Closed Set Principle**: You MUST ONLY rank the mutations provided in the list above. Do NOT introduce new mutations, insertions, deletions, or wild-type strings.
2. **Multi-Mutation Candidates**: A candidate contains multiple mutations joined by + (e.g., K23A+A40P+T52S). When ranking multi-mutation candidates, account for their combined effects.
3. **Exact Copy & Completeness**:
   - The output list MUST contain **exactly `total_candidates` items**.
   - Every candidate from the input list MUST appear **exactly once** (no duplicates, no omissions).
   - Each mutation string MUST be copied **exactly as provided**.
4. **Strict Validation**:
   - Every mutation position follows `1 <= position <= Length`.
   - `WT` MUST strictly match the character at `wildtype_sequence[position - 1]`.
   - `MUT` MUST be strictly different from `WT` (no synonymous/no-op mutations).

---

### OUTPUT FORMAT
Return a **valid JSON object ONLY** with NO markdown code block wrappers, prefix, or conversational text. Use the following exact JSON schema:

{
  "ranking": [
    "<best mutant copied exactly from the provided list>",
    "<second-best mutant copied exactly from the provided list>",
    "..."
  ]
}

---

### INSTANCE DATA
1. ASSAY CONTEXT & TARGET METRICS
- **UniProt ID**: {uniprot_id}
- **Primary Task Class**: {primary_task_class}
- **Fitness Metric Type**: {fitness_type}
- **Assay Readout Subclass**: {readout_subclass}

2. INPUT WILD-TYPE SEQUENCE (Length: {sequence_length})
`{wildtype_sequence}`

<<3. SINGLE MUTANT CONTEXT
{single_mutant_dms_scores}>>

4. CANDIDATE MUTATIONS TO RANK (Total Candidates: {num_candidates})
{candidate_mutants}
\end{promptbox}

\subsection{Confidence Elicitation for Uncertainty Estimation}
\label{app:confidence-prompt}

To elicit list-level confidence, we inserted the confidence elicitation instruction immediately before the original output contract. The original JSON ranking schema was retained, with one \texttt{confidence} field added after the ranking array; the non-numerical placeholder \texttt{<your confidence>} was used to avoid anchoring the model to an example value. The same confidence instruction was used in all settings. In \ntone, the original prohibition on ``numerical confidence scores'' was changed to ``per-mutation confidence scores'' to permit the required list-level value. All other task instructions, ranking constraints, assay information, sequences, candidate data, and contextual scores remained unchanged.

\begin{promptbox}{Confidence Elicitation Prompt Template}
You are an expert protein engineer and computational biologist specializing in deep mutational scanning (DMS) and mutation-effect prediction.

### TASK GOAL
Given a wild-type protein sequence and its experimental assay context, predict the **top 40 single point mutations** (WTposMUT format) that optimize the target fitness metric, ordered from highest expected fitness to lowest expected fitness.

### REASONING & EVIDENCE BOUNDARIES
1. **Biochemical Deductions**: Analyze residue chemistry, conservation, secondary structure propensities, steric packing, hydrophobic cores, electrostatic interactions, and sequence motifs within the supplied wild-type sequence.
2. **Assay Alignment**: Align every ranked mutation strictly with the supplied assay readout. For example, if evaluating stability/abundance, prioritize mutations that improve hydrophobic packing or thermostability without disrupting necessary structural dynamics.
3. **No Hallucination**: Do NOT invent or claim specific numerical model scores, experimental PDB coordinates, literature measurements, or alignments that are not logically derivable from sequence biochemistry.
4. **Single-Response Constraint**: You have no external tools, browsing, or follow-up turns. Complete the analysis in this single response.

### STRICT MUTATION & FORMAT CONSTRAINTS
1. **Ranking Size**: The output list MUST contain **exactly 40 mutations**, ordered from best to worst.
2. **Format**: Every mutant MUST be represented in standard 1-indexed `WTposMUT` format (e.g., `H24R`, `A15V`).
3. **Alphabet**: Both `WT` and `MUT` must be standard 20 amino acid single-letter codes: `A, C, D, E, F, G, H, I, K, L, M, N, P, Q, R, S, T, V, W, Y`.
4. **Uniqueness**: Every mutation string MUST appear exactly once (no duplicates, no omissions within your top-40 list).
5. **Strict Validation**:
   - Every mutation position follows `1 <= position <= Length`.
   - `WT` MUST strictly match the character at `wildtype_sequence[position - 1]`.
   - `MUT` MUST be strictly different from `WT` (no synonymous/no-op mutations).
6. **Search Space**: Consider all valid single amino acid substitutions across the full wild-type sequence, then return only the top 40.
7. **Prohibited Formats**: Multi-site mutations, insertions (`ins`), deletions (`del`), stop codons (`*`), HGVS notations, or per-mutation confidence scores.

---

<<### CONFIDENCE
After producing the ranking, report an overall confidence score from 0 to 100 indicating how confident you are in the predictive quality of the top-40 list for this specific protein and assay. The score should reflect confidence that the list contains and prioritizes genuinely high-fitness mutations, rather than confidence in formatting or instruction following.

Report exactly one list-level `confidence` JSON number in [0,100]. Do not add per-mutation confidence scores or any other fields. This confidence instruction must not change the ranking or item-count rules.>>

### OUTPUT FORMAT
Return a **valid JSON object ONLY** with NO markdown code block wrappers, prefix, or conversational text. Use the following exact JSON schema:

{
  "ranking": [
    "<best mutant>",
    "<second-best mutant>",
    "..."
  ],
  <<"confidence": <your confidence> >>
}

---

### INSTANCE DATA
1. ASSAY CONTEXT & TARGET METRICS
- **UniProt ID**: <uniprot_id>
- **Primary Task Class**: <primary_task_class>
- **Fitness Metric Type**: <fitness_type>
- **Assay Readout Subclass**: <assay_readout_subclass>

2. INPUT WILD-TYPE SEQUENCE (Length: <sequence_length>)
`<wildtype_sequence>`
\end{promptbox}

\clearpage
\section{Experimental Setup}

\subsection{PLM Deployment}
\label{app:protein-model-deployment}

\Cref{tab:protein-model-deployment} summarizes the model-specific inputs, checkpoints, inference configurations, and scoring definitions. Neural-network inference used FP32 precision, and all scores were oriented such that larger values indicate more favorable candidates. Each run was validated for input consistency, complete candidate coverage, finite outputs, and agreement between aggregate and component scores.

\begin{table*}[t]
\centering
\scriptsize
\renewcommand{\arraystretch}{1.12}
\begin{tabularx}{\textwidth}{@{}lXXX@{}}
\toprule
\addlinespace[0.5em]
\textbf{Model} & \textbf{Inputs and Checkpoint} & \textbf{Core Configuration} & \textbf{Candidate Score} \\
\addlinespace[0.2em]
\midrule
\addlinespace[0.5em]
ESM-2~\cite{lin2023evolutionary}
& Wild-type sequence; \texttt{esm2\_t33\_650M\_UR50D}
& Masked inference with a 1,024-token context (at most 1,022 residues); deterministic mutation-centered windows for longer chains
& Sum of wild-type-context masked-marginal log odds, $\log p(x_i^{\mathrm{mut}})-\log p(x_i^{\mathrm{wt}})$, over substituted sites \\

\addlinespace[0.2em]
ProGen2-base~\cite{nijkamp2023progen2}
& Complete wild-type and mutant sequences; \texttt{progen2-base}
& Forward and reversed causal sequence scoring with a 2,048-token context; all evaluated chains were processed at full length
& Difference between the bidirectional sequence scores of the complete mutant and wild-type chains \\

\addlinespace[0.2em]
ProSST-2048~\cite{li2024prosst}
& Wild-type sequence and AlphaFold~3 structure; \texttt{AI4Protein/ProSST-2048}
& Official GVP quantizer with a 2,048-code structural vocabulary; all evaluated chains were processed at full length within the 2,046-residue limit
& Sum of structure-conditioned wild-type-context marginal log odds over substituted sites \\

\addlinespace[0.2em]
S3F~\cite{zhang2024s3f}
& Wild-type sequence, AlphaFold~3 structure, and the corresponding molecular surface; released S3F checkpoint
& Mutation-centered windows of at most 1,022 residues; sequence logits replace structure-conditioned logits where AF3 pLDDT is below 70
& Sum of masked-marginal log-odds contributions over substituted sites \\

\addlinespace[0.2em]
VenusREM~\cite{tan2025venusrem}
& Wild-type sequence, ProSST-2048 structural tokens, and a UniRef100/MMseqs2 MSA
& \texttt{aa\_seq\_aln} retrieval; $\alpha=0.8$, sampling ratio $1.0$, and one sampling pass; full-chain inference for all evaluated contexts
& Sum of mutant-minus-wild-type values from the fused sequence, structure, and MSA representation \\

\addlinespace[0.2em]
S3F-MSA~\cite{zhang2024s3f,frazer2021eve}
& S3F score and an EVE ensemble trained on the corresponding wild-type-chain MSA
& Five EVE seeds; 400,000 optimization steps, batch size 256, learning rate $10^{-4}$; 20,000 Monte Carlo samples per seed at scoring time
& Equal average of query-wise standardized S3F and EVE scores, where the EVE score is the negative mean evolution index across seeds \\
\addlinespace[0.3em]
\bottomrule
\end{tabularx}
\caption{Deployment and scoring configurations of the protein-model baselines.}
\label{tab:protein-model-deployment}
\end{table*}

Multi-substitution scoring followed each baseline's formulation. For ESM-2, ProSST-2048, S3F, and VenusREM, substitutions on the same chain were evaluated in a shared wild-type context and their sitewise contributions were summed. ProGen2-base instead evaluated the complete mutant sequence, whereas the EVE component of S3F-MSA assigned a joint score to the complete within-chain mutation set. For candidates spanning multiple chains, each naturally mutated chain was scored separately and the chain-level contributions were summed. Thus, the protocol preserves native chain boundaries but does not explicitly model inter-chain epistasis.

The MSAs used by VenusREM and EVE were generated against UniRef100~\cite{suzek2015uniref} with MMseqs2~\cite{steinegger2017mmseqs2}. One A3M alignment was associated with each wild-type chain; lowercase insertion symbols were removed where required by the parser, while alignment gaps were retained. For EVE, the focus-column and sequence-fragment gap thresholds were 1.0 and 0.5, respectively, and the sequence-reweighting threshold was 0.01 for viral proteins and 0.2 otherwise. Five EVE models were associated with each input context. Existing EVE ensembles were reused only when the wild-type sequence, processed MSA, and reweighting configuration were identical; all benchmark mutation sets were scored anew. Structure tokens, molecular surfaces, and MSA-derived models were prepared once per wild-type context rather than for every candidate mutation.


We generated a consistent set of wild-type structures with AlphaFold~3 (AF3)~\cite{abramson2024alphafold3} for the structure-dependent baselines. Each natural chain was modeled independently as a monomer using random seed 1, 10 recycles, and five diffusion samples. The corresponding UniRef100/MMseqs2 unpaired MSA was supplied directly; templates and additional database searches were disabled. The highest-ranked sample provided the PDB input used to derive ProSST structural tokens and S3F molecular surfaces. The canonical mmCIF files, confidence outputs, sample rankings, and all five samples were retained for reproducibility.

For the benchmark release, AF3 was run for all unique wild-type chains. Each result was required to reproduce the exact target sequence and residue mapping and to contain a complete protein backbone and valid confidence arrays. Of these, 184 structures passed the primary confidence criteria; the remaining 12 had lower pTM or mean pLDDT and were retained with a review flag because they remained sequence-consistent, structurally complete, and free of detected clashes. These flags were propagated with the structural resources so that confidence-dependent analyses can be performed without changing the benchmark coverage.

\subsection{Test-time Scaling Methods}

\label{app:tts}

For each \ntaskone assay, let the $i$-th independently sampled top-40 ranking be

\begin{equation}
R_i = [m_{i1},m_{i2},\ldots,m_{i40}],
\nonumber
\end{equation}

and let $r_i(m)$ denote the one-indexed rank of mutation $m$ in $R_i$. We evaluated $n\in\{1,2,4,8,16\}$ using nested prefixes of the same 16 samples. Sample Avg was computed by evaluating each eligible raw ranking independently and averaging its assay-level metric.

All aggregation methods received the same extracted candidate strings. Candidates were deduplicated within each ranking while preserving their first occurrence. In the main scaling experiments, no explicit sequence-validity or DMS-membership filtering was applied. Incomplete rankings were padded to 40 entries using sample-specific invalid placeholders, preventing missing outputs from creating artificial agreement across samples. The aggregated top-40 output was mapped and scored only by the common evaluator.

\paragraph{Best-of-N}
We first computed the cross-sample reciprocal-rank consensus of each candidate,

\begin{equation}
c(m)=\sum_{i:m\in R_i}\frac{1}{k+r_i(m)},
\nonumber
\end{equation}

with $k=10$. Each complete ranking received the mean consensus score

\begin{equation}
q_i=\frac{1}{|R_i|}\sum_{m\in R_i}c(m),
\nonumber
\end{equation}

and the ranking with the largest $q_i$ was selected. Thus, Best-of-N always returned one intact sampled ranking and never recombined candidates across samples. Remaining ties were resolved in favor of the lower sample index.

\paragraph{Approval Voting}
Each mutation received one vote from every top-40 ranking in which it appeared,

\begin{equation}
s_{\mathrm{approval}}(m)=\sum_i \mathbb{I}[m\in R_i].
\nonumber
\end{equation}

Candidates were sorted by vote count, followed by their mean observed rank, best observed rank, and mutation string. The 40 highest-ranked candidates from the union were returned.

\paragraph{Borda Fusion}
A mutation at rank $r$ received $41-r$ points, giving

\begin{equation}
s_{\mathrm{Borda}}(m)
=
\sum_{i:m\in R_i}\left(41-r_i(m)\right).
\nonumber
\end{equation}

This score jointly reflects occurrence frequency and a linear preference for mutations placed near the top of each sampled ranking.

\paragraph{Reciprocal Rank Fusion}
RRF assigned each mutation the score

\begin{equation}
s_{\mathrm{RRF}}(m)
=
\sum_{i:m\in R_i}\frac{1}{k+r_i(m)},
\nonumber
\end{equation}

where $k=10$. Compared with Borda Fusion, reciprocal weighting places relatively greater emphasis on the highest-ranked candidates.

\paragraph{Hierarchical RRF}
To aggregate evidence shared by different substitutions at the same residue, we defined $p(m)$ as the residue position of mutation $m$ and $r_i^{\mathrm{pos}}(p)$ as the first rank at which position $p$ occurred in $R_i$. Position-level support was

\begin{equation}
s_{\mathrm{pos}}(p)
=
\sum_{i:\exists m\in R_i,\ p(m)=p}\frac{1}{k+r_i^{\mathrm{pos}}(p)},
\nonumber
\end{equation}

and the final score was

\begin{equation}
s_{\mathrm{hier}}(m)
=
s_{\mathrm{RRF}}(m)
+
\lambda\,s_{\mathrm{pos}}\!\left(p(m)\right),
\nonumber
\end{equation}

with $k=10$ and $\lambda=0.35$. Candidates whose strings could not be parsed into residue positions retained their exact-mutation RRF support but received no position-level contribution.

For Borda, RRF, and Hierarchical RRF, ties were resolved by higher occurrence frequency, better mean observed rank, and then mutation string. We used no per-position diversity cap, so multiple substitutions at the same residue could appear in the final top-40 ranking.

\clearpage
\section{Extended Experimental Analysis}

\subsection{Evaluation Completeness and Missing-Query Handling}
\label{app:llm_missing_assays}

\begin{table}[t]
\centering
\small
\renewcommand{\arraystretch}{1.05}

\begin{tabular}{l >{\rmfamily}l ccccc}

\toprule
\multirow{2}{*}{\textbf{Category}}
& \multirow{2}{*}{\textbf{Model}}
& \multicolumn{4}{c}{\textbf{Baseline}}
& \multirow{2}{*}{\textbf{Adaptive}} \\
\cmidrule(lr){3-6}
& & \textbf{T1} & \textbf{T2} & \textbf{T3} & \textbf{T4} &  \\
\midrule

\addlinespace[0.2em]
\multirow{1}{*}{\textbf{Statistic}}
  & Total              & 123 & 74 & 67 & 29 & 123 \\
\addlinespace[0.2em]

\midrule

\addlinespace[0.2em]
\multirow{6}{*}{\textbf{LLM}}
  & GPT-6 Astra        & 7 & 0 & 0 & 0 & 5 \\
  & Claude Opus 5      & 12 & 6 & 6 & 1 & 9 \\
  & Gemini 3.1 Pro     & 0 & 0 & 0 & 0 & 0 \\
  & Kimi K3            & 2 & 3 & 2 & 1 & --- \\
  & GLM-5.2            & 0 & 4 & 2 & 0 & --- \\
  & DeepSeek-V4-Pro    & 0 & 0 & 1 & 0 & 0 \\
\addlinespace[0.2em]

\midrule

\addlinespace[0.2em]
\multirow{5}{*}{\textbf{Agent}}
& Biomni $+$ GPT-6 Astra      & 9 & 1 & 1 & 1 & --- \\
& Biomni $+$ Claude Opus 5    & 0 & 4 & 1 & 0 & --- \\
\addlinespace[0.2em]
& \toolkit $+$ GPT-6 Astra      & 2 & 0 & 0 & 0 & --- \\
& \toolkit $+$ Claude Opus 5    & 1 & 2 & 0 & 1 & --- \\
& \toolkit $+$ AMix-2.1         & 0 & 0 & 0 & 1 & --- \\
\addlinespace[0.2em]

\bottomrule
\end{tabular}
\caption{Number of assay-level queries replaced by the deterministic SHA-256 random baseline during evaluation, for the baseline \textbf{\bench} setting and the multi-round adaptive search setting. A dash indicates that the corresponding task was not run for that model.}
\label{tab:llm_missing_assays}
\end{table}

Despite repeated retries, some model--task pairs involving general-purpose LLMs produced no prediction that satisfied the required output format and evaluation constraints. These failures occurred particularly when the content-filtering policies of certain general-purpose LLMs restricted specific inputs, observed most frequently for Claude Opus 5, GPT-6 Astra, Kimi K3 and GLM-5.2, with frequency varying across models and tasks. 

\paragraph{Fallback Substitutions} 
To handle this issue, we distinguished successful predictions from missing assay-level queries rather than assigning a score of zero to failed queries.
Missing queries were replaced by a deterministic SHA-256 random baseline. For each missing query, the evaluator collected all measured mutants in the corresponding ground-truth table and computed its SHA-256 digest. 
The measured mutants were ordered lexicographically by these hexadecimal digests. For \ntone, the first 40 mutants in this order were used as the substitute prediction list. For \nttwo--\ntfour, the same procedure was applied independently to each candidate query, after which the evaluator applied the corresponding task-specific ranking budget.

The numbers of substituted queries are summarized in \Cref{tab:llm_missing_assays}. Because this fallback ordering is independent of the \dmsscore ranking requirement, the reported aggregate metrics combine model predictions from successfully completed queries with the random-baseline lists for missing queries. Consequently, a larger number of missing queries will generally depress the aggregate score. 

\begin{figure}[t]
    \centering
    \includegraphics[width=1.00\linewidth]{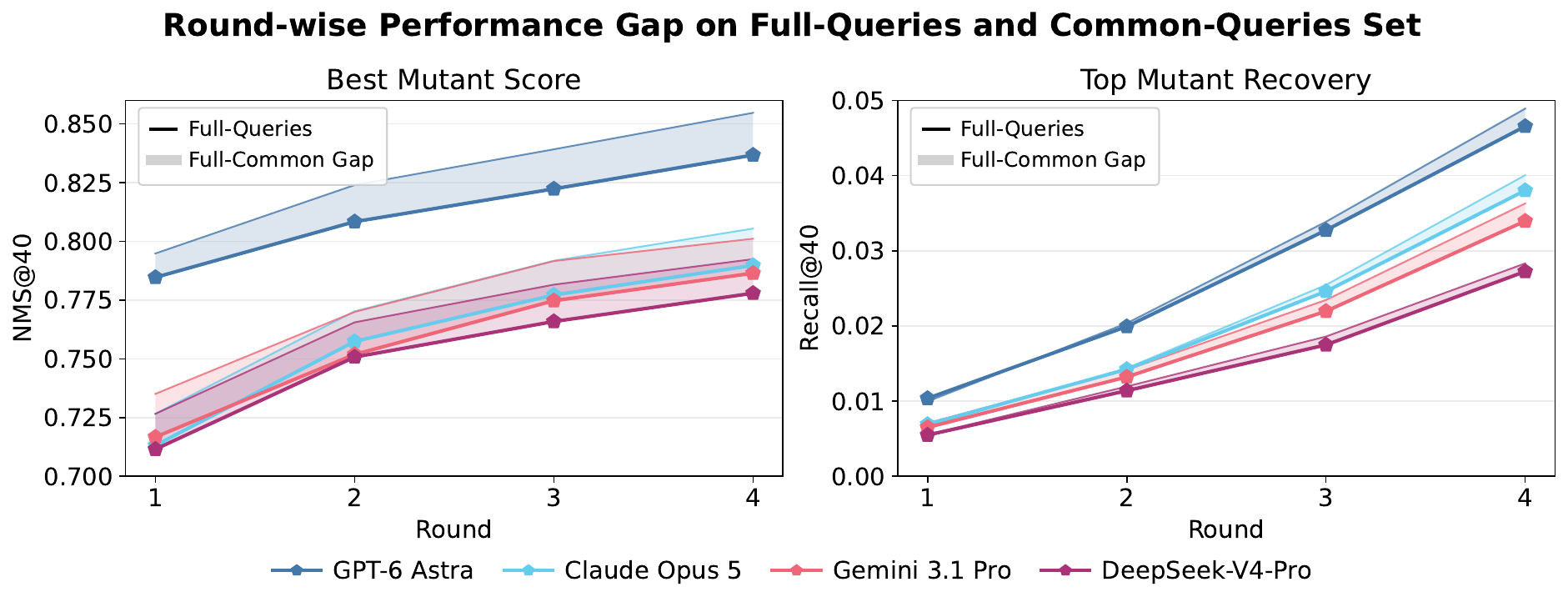}
    \caption{Matched-query analysis of round-wise single-mutant generation. Marker lines show cumulative performance on all 123 assay queries, with deterministic random-baseline substitutions used for missing queries. Thin curves show performance on the common cohort of 113 assays completed by every model in all four rounds. Shaded regions indicate the performance gap between the full and matched evaluations.}
    \label{fig:roundwise_missing_assays}
\end{figure}

\paragraph{Matched-query Analysis}
To isolate intrinsic model performance from the effects of incomplete outputs, we curated the multi-round adaptive search evaluation of \Cref{sec:multiround_adaptation} on all 123 ground-truth assay queries to a common cohort of 113 assays, as shown in \Cref{fig:roundwise_missing_assays}. The common cohort contained only assays for which every model produced a valid cumulative prediction prefix across all four rounds; no random substitution was used for these assays. By contrast, the original multi-round evaluation retained all 123 assays and used the deterministic random fallback whenever a model lacked a complete prediction. At the fourth round, complete-prefix coverage ranged from 114 assays for Claude Opus 5 and 118 assays for GPT-6 Astra to all complete 123 assays for Gemini 3.1 Pro and DeepSeek-V4-Pro.

Across the common cohort, all models showed progressive improvements in both NMS@$40$ and Recall@$40$ over successive rounds. At round four, common-cohort NMS@$40$ exceeded the corresponding full-cohort value by $0.018$ for GPT-6 Astra and by approximately $0.015$ for the other three models. Recall@$40$ was boosted by approximately $0.0011$ to $0.0023$ across models. The higher common-cohort scores for Gemini 3.1 Pro and DeepSeek-V4-Pro, which had complete coverage of all 123 assays, suggest that the 10 assays excluded from the matched cohort due to incomplete predictions from other models were more difficult on average.
These common-cohort results therefore provide a cleaner comparison of intrinsic model quality, whereas the full-cohort curves preserve performance under the original evaluation protocol.

\subsection{Complementary Strengths and Failure Modes of LLMs and PLMs}
\label{subsec:plmvsllm-revised}

We tested whether PLMs and LLMs identify the same useful mutants and make the same ranking errors. GPT-6 Astra represented the LLM family, and VenusREM represented the PLM family. The comparison had two parts: \ntone measured overlap and experimental quality among generated single-mutant candidates, whereas \nttwo--\ntfour measured whether one model could correct extreme ranking errors made by the other. This design separates complementary exploration in \ntone from complementary error correction in multi-mutant ranking.

\begin{figure}[t]
    \centering
    \includegraphics[width=1.00\linewidth]{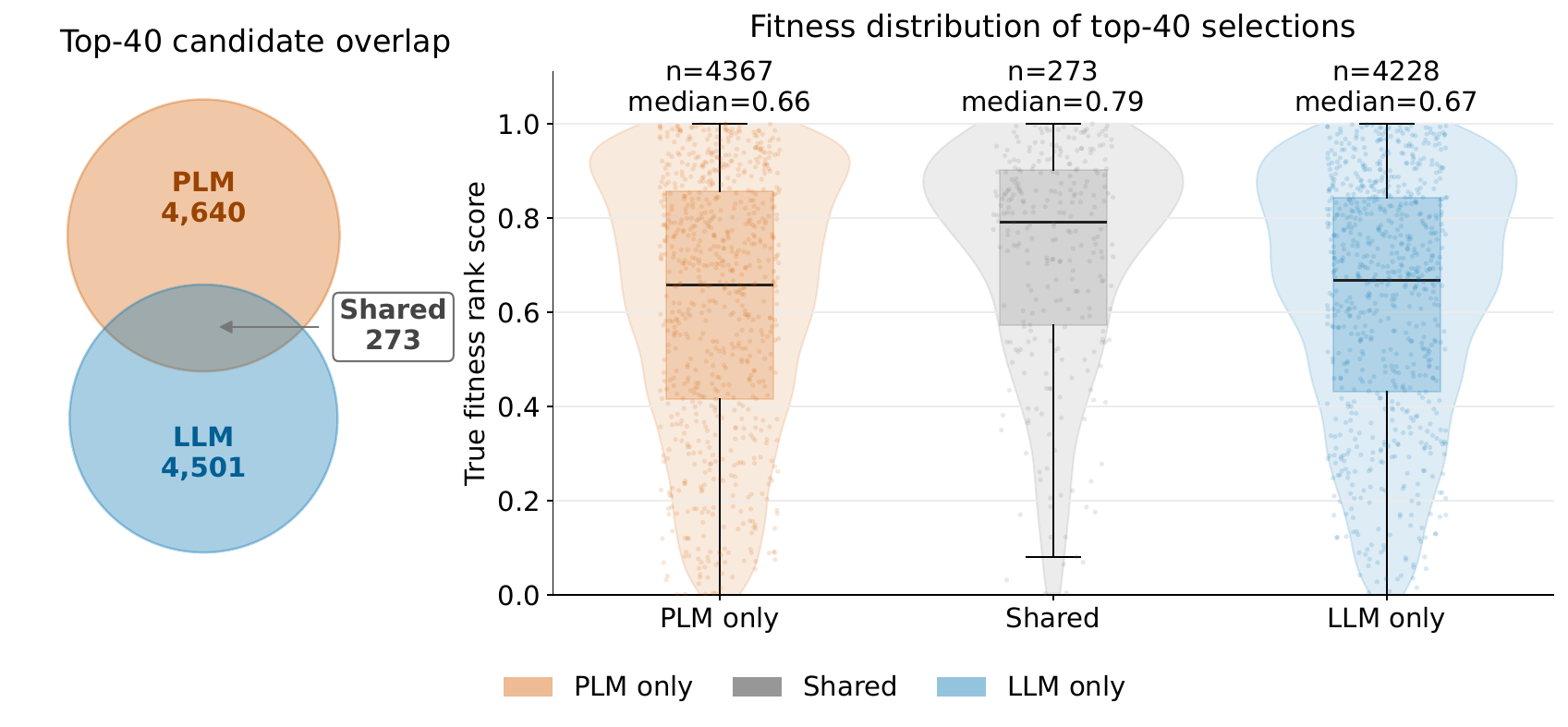}
    \caption{\textbf{Complementary single-mutant selection in \ntone.} Left, overlap between evaluable top-$40$ selections from VenusREM (PLM) and GPT-6 Astra (LLM) across 116 paired assays. Right, experimental fitness-rank distributions of PLM-only, shared and LLM-only selections. Rank scores were normalized within each assay from 1 (best) to 0 (worst). n denotes mutant–assay selections.}
    \label{fig:t1_between_llm_plm_revised}
\end{figure}

\paragraph{T1: Single-mutant Exploration} 
GPT-6 Astra and VenusREM selected largely different single-mutant candidates, but their small consensus set was the most reliable. As shown in \Cref{fig:t1_between_llm_plm_revised}, across 116 paired assays, VenusREM and
GPT-6 Astra produced 4,640 and 4,501 evaluable top-$40$ selections, respectively, with 273 shared mutant--assay pairs. 
The shared candidates therefore represented only a small fraction of the selections made by either model. 
Nevertheless, the consensus set had the highest experimental quality: the median true rank score was 0.79 for shared candidates, compared with 0.66 for VenusREM-only candidates and 0.67 for GPT-6 Astra-only candidates, where 1 denotes the best experimental rank. Thus, agreement was uncommon but informative: consensus candidates can provide a high-confidence shortlist, whereas model-specific candidates expand the explored sequence space.

\begin{figure}[!htbp]
    \centering
    \includegraphics[width=1.00\linewidth]{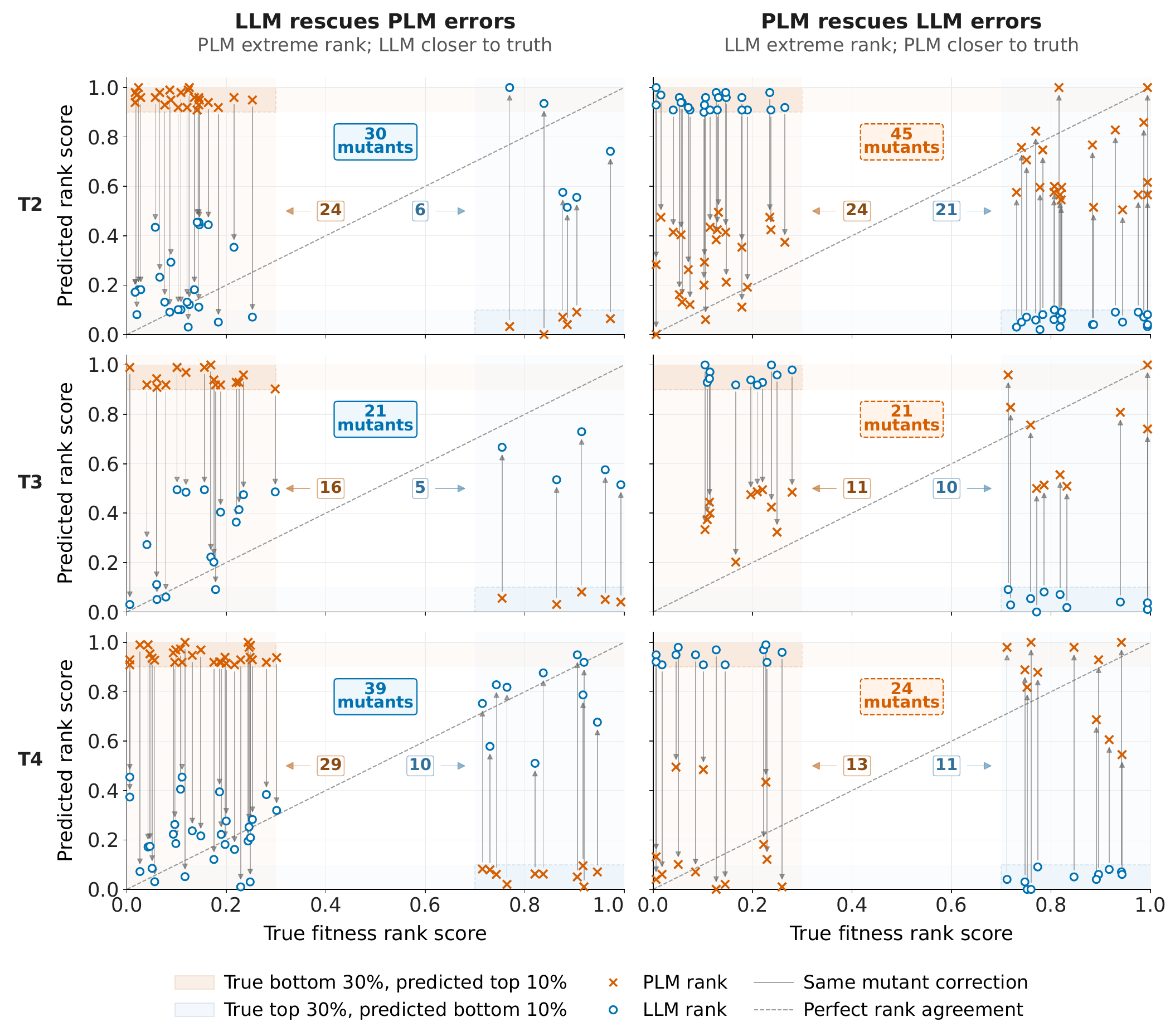}
    \caption{\textbf{Complementary correction of extreme ranking errors by GPT-6 Astra and VenusREM across \nttwo–\ntfour.} The left column shows mutants for which GPT-6 Astra corrected extreme VenusREM ranking errors, and the right column shows the converse. True fitness rank scores range from 0 (worst) to 1 (best), whereas predicted rank scores range from 0 (bottom) to 1 (top). Orange crosses and blue circles denote VenusREM and GPT-6 Astra predictions, respectively; grey arrows connect predictions for the same mutant. Shaded regions indicate true bottom-30\% mutants predicted in the top 10\% or true top-30\% mutants predicted in the bottom 10\%. Central labels report the total number of mutants in each direction.}
    \label{fig:extreme_correction_revised}
\end{figure}

\paragraph{T2--T4: Ranking Errors} 
We next examined whether LLMs and PLMs make complementary ranking errors in the multi-mutant tasks. For each task, experimental fitness and model predictions were converted into within-assay rank scores, with 1 denoting the best mutant. In \Cref{fig:extreme_correction_revised}, the $x$-axis shows the true fitness rank score and the $y$-axis shows the predicted rank score. The diagonal corresponds to perfect rank agreement, while vertical arrows connect the two predictions for the same mutant. We defined an extreme error as placing a true bottom-30\% mutant in the predicted top 10\% or a true top-30\% mutant in the predicted bottom 10\%. A correction was counted when the other model assigned the same mutant a substantially less extreme rank, moving it closer to its measured rank. 
This paired, case-level analysis distinguishes shared failures from model-specific errors that can be rescued by the other model.
The direction-specific counts further show that the corrected cases include both low-fitness over-rankings and high-fitness under-rankings, indicating that the complementarity is not restricted to a single type of ranking failure.

The error-correction patterns in \nttwo and \ntthree indicate that VenusREM remains at least as reliable as GPT-6 Astra when limited or localized experimental evidence is available, while they still provide complementary predictions. In \nttwo, VenusREM corrected 45 extreme GPT-6 Astra errors, including 24 low-fitness over-rankings and 21 high-fitness under-rankings, whereas GPT-6 Astra corrected 30 VenusREM errors, including 24 and 6 of these two types, respectively. In \ntthree, the two models corrected the same total number of extreme errors, with GPT-6 Astra correcting 16 low-fitness over-rankings and 5 high-fitness under-rankings, and VenusREM correcting 11 and 10, respectively.  

The availability of complete single-mutant context reverses the direction of error correction in favor of GPT-6 Astra. In \ntfour, GPT-6 Astra corrected 39 extreme VenusREM errors, including 29 low-fitness over-rankings and 10 high-fitness under-rankings, whereas VenusREM corrected 24 GPT-6 Astra errors, including 13 and 11 of these two types, respectively. This reversal mirrors the higher aggregate performance of LLM-based systems in \ntfour, 
where measured fitness values for the component single mutations are available as additional evidence for ranking multi-mutant candidates. The results therefore support a context-dependent shift in model utility rather than a universal superiority of either model family. 

Taken together, VenusREM and GPT-6 Astra exhibit complementary ranking errors across \nttwo--\ntfour, with each correcting both low-fitness over-rankings and high-fitness under-rankings made by the other. These patterns motivate combined or agent-mediated approaches, but this descriptive analysis does not establish that an ensemble would improve aggregate performance.

%% file: refs.bib
@article{qiu2026amix,
  title={AMix-2: Establishing Protein as a Native Modality in Large Language Models},
  author={Qiu, Keyue and Wu, Yixin and Wang, Lihao and Ouyang, Yawen and Yu, Jixiang and Zhou, Zihan and Lv, Changze and Xue, Dongyu and Song, Yuxuan and Zhang, Xinbo and others},
  journal={arXiv preprint arXiv:2605.30963},
  year={2026}
}

@inproceedings{
wang2026towards,
title={Towards A Generative Protein Evolution Machine with {DPLM}-Evo},
author={Xinyou Wang and Liang Hong and Jiasheng Ye and Zaixiang Zheng and Shujian Huang and Quanquan Gu},
booktitle={ICLR 2026 Workshop on Generative and Experimental Perspectives for Biomolecular Design},
year={2026},
url={https://openreview.net/forum?id=xqnTGzQmfB}
}

@misc{openai2026gpt6astra,
  title={GPT-6 Astra: A new generation of intelligence},
  author={Openai},
  year={2026},
  howpublished={\url{https://openai.com/index/gpt-6-astra/}}
}

@misc{anthropic2026claude48,
  author       = {{Anthropic}},
  title        = {Claude Opus 4.8 System Card},
  year         = {2026},
  month        = may,
  howpublished = {\url{https://www-cdn.anthropic.com/0b4915911bb0d19eca5b5ee635c80fef830a37ea.pdf}}
}

@misc{anthropic2026claude5,
  title={Introducing Claude Opus 5},
  author={Anthropic},
  year={2026},
  howpublished={\url{https://www.anthropic.com/news/claude-opus-5}}
}

@misc{google2026gemini3,
  title={Gemini 3.1 Pro},
  author={Google DeepMind},
  year={2026},
  howpublished={\url{https://deepmind.google/models/gemini/pro}}
}

@misc{kimiteam2026kimik3openfrontier,
      title={Kimi K3: Open Frontier Intelligence}, 
      author={Kimi Team},
      year={2026},
      eprint={2607.24653},
      archivePrefix={arXiv},
      primaryClass={cs.CL},
      url={https://arxiv.org/abs/2607.24653}, 
}

@misc{zai2026glm52,
  title={GLM-5.2: Built for Long-Horizon Tasks},
  author={Z.ai},
  year={2026},
  howpublished = {\url{https://z.ai/blog/glm-5.2}},
}

@misc{deepseek2026v4,
  title={DeepSeek-V4: Towards Highly Efficient Million-Token Context Intelligence},
  author={DeepSeek AI},
  year={2026},
  howpublished={\url{https://huggingface.co/deepseek-ai/DeepSeek-V4-Pro}}
}

@inproceedings{dallago2021flip,
  title = {{FLIP}: Benchmark Tasks in Fitness Landscape Inference for Proteins},
  author = {Dallago, Christian and Mou, Jody and Johnston, Kadina E. and Wittmann, Bruce J. and Bhattacharya, Nicholas and Goldman, Samuel and Madani, Ali and Yang, Kevin K.},
  booktitle = {Neural Information Processing Systems Datasets and Benchmarks Track},
  year = {2021},
  url = {https://datasets-benchmarks-proceedings.neurips.cc/paper/2021/hash/2b44928ae11fb9384c4cf38708677c48-Abstract-round2.html}
}

@inproceedings{didi2026flip2,
  title = {{FLIP2}: Expanding Protein Fitness Landscape Benchmarks for Real-World Machine Learning Applications},
  author = {Didi, Kieran and Alamdari, Sarah and Lu, Alex X. and Wittmann, Bruce and Johnston, Kadina E. and Amini, Ava A. and Madani, Ali and Czeneszew, Maya and Dallago, Christian and Yang, Kevin K.},
  booktitle = {Forty-third International Conference on Machine Learning},
  year = {2026},
  url = {https://flip.protein.properties/}
}

@inproceedings{notin2023proteingym,
  title = {{ProteinGym}: Large-Scale Benchmarks for Protein Fitness Prediction and Design},
  author = {Notin, Pascal and Kollasch, Aaron and Ritter, Daniel and van Niekerk, Lood and Paul, Steffanie and Spinner, Han and Rollins, Nathan and Shaw, Ada and Orenbuch, Rose and Weitzman, Ruben and Frazer, Jonathan and Dias, Mafalda and Franceschi, Dinko and Gal, Yarin and Marks, Debora S.},
  booktitle = {Advances in Neural Information Processing Systems},
  year = {2023},
  url = {https://mlanthology.org/neurips/2023/notin2023neurips-proteingym/}
}

@article{esposito2019mavedb,
  title = {{MaveDB}: an open-source platform to distribute and interpret data from multiplexed assays of variant effect},
  author = {Esposito, Daniel and Weile, Jochen and Shendure, Jay and Starita, Lea M. and Papenfuss, Anthony T. and Roth, Frederick P. and Fowler, Douglas M. and Rubin, Alan F.},
  journal = {Genome Biology},
  volume = {20},
  number = {1},
  pages = {223},
  year = {2019},
  doi = {10.1186/s13059-019-1845-6},
  url = {https://doi.org/10.1186/s13059-019-1845-6}
}

@article{arora2026proteingymllm,
  title = {{PG-LLM}: Benchmarking General-Purpose Language Models for Protein Variant Ranking},
  author = {Arora, Rohit Krishan and Chen, Leo Tianlai and Du, Melissa and Marks, Debora and Church, George},
  journal = {bioRxiv},
  year = {2026},
  doi = {10.64898/2026.07.27.741045},
  url = {https://www.biorxiv.org/content/10.64898/2026.07.27.741045v1}
}

@misc{kim2026biodesignbench,
  title = {Evaluating {LLM}-Driven Protein Design: Agents Lack Iterative Evaluation Depth},
  author = {Kim, Jeonghyeon and Romero, Philip},
  year = {2026},
  howpublished = {GitHub repository},
  url = {https://github.com/RomeroLab/BioDesignBench},
  note = {BioDesignBench}
}

@article{lin2023evolutionary,
  title = {Evolutionary-scale prediction of atomic-level protein structure with a language model},
  author = {Lin, Zeming and Akin, Halil and Rao, Roshan and Hie, Brian and Zhu, Zhongkai and Lu, Wenting and Smetanin, Nikita and Verkuil, Robert and Kabeli, Ori and Shmueli, Yaniv and dos Santos Costa, Allan and Fazel-Zarandi, Maryam and Sercu, Tom and Candido, Salvatore and Rives, Alexander},
  journal = {Science},
  volume = {379},
  number = {6637},
  pages = {1123--1130},
  year = {2023},
  month = mar,
  publisher = {American Association for the Advancement of Science (AAAS)},
  doi = {10.1126/science.ade2574},
  url = {https://doi.org/10.1126/science.ade2574}
}

@article{nijkamp2023progen2,
  title = {{ProGen2}: Exploring the boundaries of protein language models},
  author = {Nijkamp, Erik and Ruffolo, Jeffrey A. and Weinstein, Eli N. and Naik, Nikhil and Madani, Ali},
  journal = {Cell Systems},
  volume = {14},
  number = {11},
  pages = {968--978.e3},
  year = {2023},
  month = nov,
  publisher = {Elsevier BV},
  doi = {10.1016/j.cels.2023.10.002},
  url = {https://doi.org/10.1016/j.cels.2023.10.002}
}

@inproceedings{li2024prosst,
  title = {{ProSST}: Protein Language Modeling with Quantized Structure and Disentangled Attention},
  author = {Li, Mingchen and Tan, Yang and Ma, Xinzhu and Zhong, Bozitao and Yu, Huiqun and Zhou, Ziyi and Ouyang, Wanli and Zhou, Bingxin and Tan, Pan and Hong, Liang},
  booktitle = {Advances in Neural Information Processing Systems},
  volume = {37},
  pages = {35700--35726},
  year = {2024},
  editor = {A. Globerson and L. Mackey and D. Belgrave and A. Fan and U. Paquet and J. Tomczak and C. Zhang},
  publisher = {Curran Associates, Inc.},
  doi = {10.52202/079017-1126},
  url = {https://proceedings.neurips.cc/paper_files/paper/2024/file/3ed57b293db0aab7cc30c44f45262348-Paper-Conference.pdf}
}

@inproceedings{zhang2024s3f,
  title = {Multi-Scale Representation Learning for Protein Fitness Prediction},
  author = {Zhang, Zuobai and Notin, Pascal and Huang, Yining and Lozano, Aur\'{e}lie and Chenthamarakshan, Vijil and Marks, Debora and Das, Payel and Tang, Jian},
  booktitle = {Advances in Neural Information Processing Systems},
  volume = {37},
  pages = {101456--101473},
  year = {2024},
  editor = {A. Globerson and L. Mackey and D. Belgrave and A. Fan and U. Paquet and J. Tomczak and C. Zhang},
  publisher = {Curran Associates, Inc.},
  doi = {10.52202/079017-3217},
  url = {https://proceedings.neurips.cc/paper_files/paper/2024/file/b7d795e655c1463d7299688d489e8ef4-Paper-Conference.pdf}
}

@article{frazer2021eve,
  title = {Disease variant prediction with deep generative models of evolutionary data},
  author = {Frazer, Jonathan and Notin, Pascal and Dias, Mafalda and Gomez, Aidan and Min, Joseph K. and Brock, Kelly and Gal, Yarin and Marks, Debora S.},
  journal = {Nature},
  volume = {599},
  number = {7883},
  pages = {91--95},
  year = {2021},
  month = oct,
  publisher = {Springer Science and Business Media LLC},
  doi = {10.1038/s41586-021-04043-8},
  url = {https://doi.org/10.1038/s41586-021-04043-8}
}

@article{tan2025venusrem,
  title = {From high-throughput evaluation to wet-lab studies: advancing mutation effect prediction with a retrieval-enhanced model},
  author = {Tan, Yang and Wang, Ruilin and Wu, Banghao and Hong, Liang and Zhou, Bingxin},
  journal = {Bioinformatics},
  volume = {41},
  number = {Supplement\_1},
  pages = {i401--i409},
  year = {2025},
  month = jul,
  publisher = {Oxford University Press (OUP)},
  doi = {10.1093/bioinformatics/btaf189},
  url = {https://doi.org/10.1093/bioinformatics/btaf189}
}

@article{abramson2024alphafold3,
  title = {Accurate structure prediction of biomolecular interactions with {AlphaFold} 3},
  author = {Abramson, Josh and Adler, Jonas and Dunger, Jack and Evans, Richard and Green, Tim and Pritzel, Alexander and Ronneberger, Olaf and Willmore, Lindsay and Ballard, Andrew J. and Bambrick, Joshua and Bodenstein, Sebastian W. and Evans, David A. and Hung, Chia-Chun and O'Neill, Michael and Reiman, David and Tunyasuvunakool, Kathryn and Wu, Zachary and {\v{Z}}emgulyt{\.e}, Akvil{\.e} and Arvaniti, Eirini and Beattie, Charles and Bertolli, Ottavia and Bridgland, Alex and Cherepanov, Alexey and Congreve, Miles and Cowen-Rivers, Alexander I. and Cowie, Andrew and Figurnov, Michael and Fuchs, Fabian B. and Gladman, Hannah and Jain, Rishub and Khan, Yousuf A. and Low, Caroline M. R. and Perlin, Kuba and Potapenko, Anna and Savy, Pascal and Singh, Sukhdeep and Stecula, Adrian and Thillaisundaram, Ashok and Tong, Catherine and Yakneen, Sergei and Zhong, Ellen D. and Zielinski, Michal and {\v{Z}}{\'i}dek, Augustin and Bapst, Victor and Kohli, Pushmeet and Jaderberg, Max and Hassabis, Demis and Jumper, John M.},
  journal = {Nature},
  volume = {630},
  number = {8016},
  pages = {493--500},
  year = {2024},
  month = may,
  publisher = {Springer Science and Business Media LLC},
  doi = {10.1038/s41586-024-07487-w},
  url = {https://doi.org/10.1038/s41586-024-07487-w}
}

@article{steinegger2017mmseqs2,
  title = {{MMseqs2} enables sensitive protein sequence searching for the analysis of massive data sets},
  author = {Steinegger, Martin and S{\"o}ding, Johannes},
  journal = {Nature Biotechnology},
  volume = {35},
  number = {11},
  pages = {1026--1028},
  year = {2017},
  month = oct,
  publisher = {Springer Science and Business Media LLC},
  doi = {10.1038/nbt.3988},
  url = {https://doi.org/10.1038/nbt.3988}
}

@article{suzek2015uniref,
  title = {{UniRef} clusters: a comprehensive and scalable alternative for improving sequence similarity searches},
  author = {Suzek, Baris E. and Wang, Yuqi and Huang, Hongzhan and McGarvey, Peter B. and Wu, Cathy H. and {UniProt Consortium}},
  journal = {Bioinformatics},
  volume = {31},
  number = {6},
  pages = {926--932},
  year = {2015},
  month = mar,
  publisher = {Oxford University Press (OUP)},
  doi = {10.1093/bioinformatics/btu739},
  url = {https://doi.org/10.1093/bioinformatics/btu739}
}

@article{biswas2021lown,
  author  = {Biswas, Surojit and Khimulya, Grigory and Alley, Ethan C. and Esvelt, Kevin M. and Church, George M.},
  title   = {Low-{N} protein engineering with data-efficient deep learning},
  journal = {Nature Methods},
  year    = {2021},
  volume  = {18},
  number  = {4},
  pages   = {389--396},
  doi     = {10.1038/s41592-021-01100-y},
  url     = {https://doi.org/10.1038/s41592-021-01100-y}
}

@article{hsu2022fitness,
  author  = {Hsu, Chloe and Nisonoff, Hunter and Fannjiang, Clara and Listgarten, Jennifer},
  title   = {Learning protein fitness models from evolutionary and assay-labeled data},
  journal = {Nature Biotechnology},
  year    = {2022},
  volume  = {40},
  number  = {7},
  pages   = {1114--1122},
  doi     = {10.1038/s41587-021-01146-5},
  url     = {https://doi.org/10.1038/s41587-021-01146-5}
}

@article{hopf2017evmutation,
  author  = {Hopf, Thomas A. and Ingraham, John B. and Poelwijk, Frank J. and Sch{\"a}rfe, Charlotta P. I. and Springer, Michael and Sander, Chris and Marks, Debora S.},
  title   = {Mutation effects predicted from sequence co-variation},
  journal = {Nature Biotechnology},
  year    = {2017},
  volume  = {35},
  number  = {2},
  pages   = {128--135},
  doi     = {10.1038/nbt.3769},
  url     = {https://doi.org/10.1038/nbt.3769}
}

@article{riesselman2018deepsequence,
  author  = {Riesselman, Adam J. and Ingraham, John B. and Marks, Debora S.},
  title   = {Deep generative models of genetic variation capture the effects of mutations},
  journal = {Nature Methods},
  year    = {2018},
  volume  = {15},
  number  = {10},
  pages   = {816--822},
  doi     = {10.1038/s41592-018-0138-4},
  url     = {https://doi.org/10.1038/s41592-018-0138-4}
}

@inproceedings{meier2021zeroshot,
  author    = {Meier, Joshua and Rao, Roshan and Verkuil, Robert and Liu, Jason and Sercu, Tom and Rives, Alexander},
  title     = {Language models enable zero-shot prediction of the effects of mutations on protein function},
  booktitle = {Advances in Neural Information Processing Systems},
  year      = {2021},
  volume    = {34},
  pages     = {29287--29303},
  url       = {https://proceedings.neurips.cc/paper/2021/hash/f51338d736f95dd42427296047067694-Abstract.html}
}

@inproceedings{notin2022tranception,
  author    = {Notin, Pascal and Dias, Mafalda and Frazer, Jonathan and Marchena-Hurtado, Javier and Gomez, Aidan N. and Marks, Debora S. and Gal, Yarin},
  title     = {{Tranception}: Protein fitness prediction with autoregressive transformers and inference-time retrieval},
  booktitle = {Proceedings of the 39th International Conference on Machine Learning},
  year      = {2022},
  volume    = {162},
  series    = {Proceedings of Machine Learning Research},
  pages     = {16990--17017},
  publisher = {PMLR},
  url       = {https://proceedings.mlr.press/v162/notin22a.html}
}

@article{huang2025biomni,
  author  = {Huang, Kexin and Zhang, Serena and Wang, Hanchen and Qu, Yuanhao and Lu, Yingzhou and Roohani, Yusuf and Li, Ryan and Qiu, Lin and Zhang, Junze and Di, Yin and others},
  title   = {{Biomni}: A general-purpose biomedical {AI} agent},
  journal = {bioRxiv},
  year    = {2025},
  eprint  = {2025.05.30.656746},
  doi     = {10.1101/2025.05.30.656746},
  url     = {https://doi.org/10.1101/2025.05.30.656746},
  note    = {Preprint}
}

@inproceedings{wang2025llmoptimizer,
    title     = {Large Language Model is Secretly a Protein Sequence Optimizer},
    author    = {Wang, Yinkai and He, Jiaxing and Du, Yuanqi and Chen, Xiaohui and Li, Jianan Canal and Liu, Li-Ping and Xu, Xiaolin
    and Hassoun, Soha},
    booktitle = {ICLR 2025 Workshop on Learning Meaningful Representations of Life},
    year      = {2025},
    eprint    = {2501.09274},
    archivePrefix = {arXiv},
    primaryClass  = {q-bio.BM},
    url       = {https://arxiv.org/abs/2501.09274}
  }

@article{ghafarollahi2024protagents,
    title   = {{ProtAgents}: Protein Discovery via Large Language Model Multi-Agent Collaborations Combining Physics and Machine
    Learning},
    author  = {Ghafarollahi, Alireza and Buehler, Markus J.},
    journal = {Digital Discovery},
    volume  = {3},
    number  = {7},
    pages   = {1389--1409},
    year    = {2024},
    doi     = {10.1039/D4DD00013G},
    url     = {https://doi.org/10.1039/D4DD00013G},
    publisher = {Royal Society of Chemistry}
  }

@article{tsuboyama2023megascale,
  author = {Tsuboyama, Kotaro and Dauparas, Justas and Chen, Jonathan and Laine, Elodie and Mohseni Behbahani, Yasser and Weinstein, Jonathan J. and Mangan, Niall M. and Ovchinnikov, Sergey and Rocklin, Gabriel J.},
  title = {Mega-scale experimental analysis of protein folding stability in biology and design},
  journal = {Nature},
  volume = {620},
  pages = {434--444},
  year = {2023},
  doi = {10.1038/s41586-023-06328-6}
}

@article{chungyoun2024flab,
  author = {Chungyoun, Michael and Ruffolo, Jeff and Gray, Jeffrey J.},
  title = {FLAb: Benchmarking tasks in fitness landscape inference for antibodies},
  journal = {bioRxiv},
  year = {2024},
  doi = {10.1101/2024.01.13.575504}
}

@article{beltran2025domainome,
  author = {Beltran, Antoni and Jiang, Xiang'er and Shen, Yue and Lehner, Ben},
  title = {Site-saturation mutagenesis of 500 human protein domains},
  journal = {Nature},
  volume = {637},
  number = {8047},
  pages = {885--894},
  year = {2025},
  doi = {10.1038/s41586-024-08370-4}
}

@article{chen2026combingym,
  author = {Chen, Yongcan and Fu, Lihao and Lu, Xuchao and Li, Wenzhuo and Gao, Yuan and Wang, Yibo and Ruan, Zhicheng and Si, Tong},
  title = {CombinGym: a benchmark platform for machine learning-assisted design of combinatorial protein variants},
  journal = {bioRxiv},
  year = {2026},
  doi = {10.64898/2026.03.24.714074}
}

@article{kimura2024cdkn2a,
  author = {Kimura, Hirokazu and Lahouel, Kamel and Tomasetti, Cristian and Roberts, Nicholas Jason},
  title = {Functional characterization of all CDKN2A missense variants and comparison to in silico models of pathogenicity},
  journal = {eLife},
  volume = {13},
  pages = {RP95347},
  year = {2024},
  doi = {10.7554/eLife.95347}
}

@article{wang2025slc13a5,
  author = {Wang, Wen-An and Ferrada, Evandro and Klimek, Christoph and Osthushenrich, Tanja and MacNamara, Aidan and Wiedmer, Tabea and Superti-Furga, Giulio},
  title = {Large-scale experimental assessment of variant effects on the structure and function of the citrate transporter SLC13A5},
  journal = {Science Advances},
  volume = {11},
  number = {26},
  pages = {eadx3011},
  year = {2025},
  doi = {10.1126/sciadv.adx3011}
}

@article{johnston2024trpb,
  author = {Johnston, Kadina E. and Almhjell, Patrick J. and Watkins-Dulaney, Ella J. and Liu, Grace and Porter, Nicholas J. and Yang, Jason and Arnold, Frances H.},
  title = {A combinatorially complete epistatic fitness landscape in an enzyme active site},
  journal = {Proceedings of the National Academy of Sciences},
  volume = {121},
  number = {32},
  pages = {e2400439121},
  year = {2024},
  doi = {10.1073/pnas.2400439121}
}

@article{robien2004improved,
    author  = {Robien, Mark A. and Nguyen, Kiet T. and Kumar, Abhinav and
               Hirsh, Irwin and Turley, Stewart and Pei, Dehua and
               Hol, Wim G. J.},
    title   = {An improved crystal form of {Plasmodium falciparum}
               peptide deformylase},
    journal = {Protein Science},
    year    = {2004},
    volume  = {13},
    number  = {4},
    pages   = {1155--1163},
    doi     = {10.1110/ps.03456404}
}
